\documentclass[nohyperref]{article}

\usepackage{microtype}
\usepackage{graphicx}
\usepackage{subfigure}
\usepackage{booktabs} 

\usepackage{hyperref}

\usepackage[accepted]{icml2022}

\usepackage{amsmath}
\usepackage{amssymb}
\usepackage{mathtools}
\usepackage{amsthm}

\usepackage[capitalize,noabbrev]{cleveref}

\usepackage{amssymb,amsmath,amsthm,dsfont,bm,mathtools,amsfonts,mathabx}
\usepackage{enumitem} 
\setlist{topsep=10pt, partopsep=0pt, leftmargin=13pt} 
\usepackage{xcolor,color,graphicx,epsfig}
\usepackage[multiple]{footmisc}
\usepackage{nicefrac}

\newcommand{\Loss}{{\mathcal{L}}}
\newcommand{\loss}{{L}}

\newcommand{\Lossapx}{{\hat{\Loss}}}
\newcommand{\tmplt}[3]{{{#1}_{\mathrm{#2}}^{\mathtt{#3}}}}
\newcommand{\smplst}{{\mathcal{S}}}

\newcommand{\dec}{{\Delta}}

\newcommand{\decsc}{\dec^{\mathtt{SC}}}
\newcommand{\decgsc}{\dec^{\mathtt{GSC}}}

\newcommand{\decgp}{\dec^{\mathtt{GP}}}

\newcommand{\decnl}{\dec^{\mathtt{NL}}}

\newcommand{\decppe}{\dec^{\mathtt{PPE}}}
\newcommand{\decppetilde}{\tilde{\dec}^{\mathtt{PPE}}}
\newcommand{\util}{{u}}
\newcommand{\putil}{{\tilde{\util}}}

\newcommand{\yhat}{{\hat{y}}}
\newcommand{\ytilde}{{\tilde{y}}}

\newcommand{\what}{{\hat{w}}}
\newcommand{\wbar}{{\widebar{w}}}

\newcommand{\flipcost}{{\phi}}
\newcommand{\canflip}{{\Phi}}
\newcommand{\margin}{{\gamma}}

\newcommand{\err}{{\texttt{err}}} 

\newcommand{\beq}{\begin{equation}}
\newcommand{\eeq}{\end{equation}}
\newcommand{\bal}{\begin{align}}
\newcommand{\eal}{\end{align}}
\newcommand\expect[2]{\mathbb{E}_{#1}{[ {#2} ]}}
\newcommand\prob[2]{\mathbb{P}_{#1}{\left[ {#2} \right]}}
\DeclareMathOperator*{\argmax}{argmax}
\DeclareMathOperator*{\argmin}{argmin}

\DeclareMathOperator{\sign}{sign}

\newcommand\norm[1]{\left\| {#1} \right\|}

\newcommand{\1}[1]{\mathds{1}{\{{#1}\}}}

\newcommand{\naive}{{na\"{\i}ve}}

\newcommand{\X}{{\cal{X}}}

\newcommand{\Y}{{\cal{Y}}}

\newcommand{\Z}{{\cal{Z}}}

\renewcommand{\H}{{\cal{H}}}

\newcommand{\R}{{\mathbb{R}}}

\theoremstyle{plain}
\newtheorem{theorem}{Theorem} 

\newtheorem{lemma}{Lemma}
\newtheorem{corollary}{Corollary}
\theoremstyle{definition}
\newtheorem{definition}{Definition}

\theoremstyle{remark}

\begin{document}

\twocolumn[
\icmltitle{Generalized Strategic Classification and the Case of Aligned Incentives}




\begin{icmlauthorlist}
\icmlauthor{Sagi Levanon}{CS}
\icmlauthor{Nir Rosenfeld}{CS}
\end{icmlauthorlist}

\icmlaffiliation{CS}{Faculty of Computer Science, Technion - Israel Institute of Technology, Haifa, Israel}

\icmlcorrespondingauthor{Nir Rosenfeld}{nirr@cs.technion.ac.il}

\icmlkeywords{Machine Learning, ICML}

\vskip 0.3in
]



\printAffiliationsAndNotice{}  

\begin{abstract}
Strategic classification studies learning
in settings where self-interested users
can strategically modify their features to
obtain favorable predictive outcomes.
A key working assumption, however, is that
``favorable'' always means ``positive'';
this may be appropriate in some applications
(e.g., loan approval),
but reduces to a fairly narrow view of what user interests can be.
In this work we argue for a broader perspective
on what accounts for strategic user behavior,
and propose and study a flexible model of
\emph{generalized strategic classification}.
Our generalized model subsumes most current models,
but includes other novel settings;
among these, we identify and target one intriguing sub-class of problems in which the interests of users and the system are \emph{aligned}.
This setting reveals a surprising fact:
that standard max-margin losses are ill-suited for strategic inputs.
Returning to our fully generalized model,
we propose a novel max-margin framework for \emph{strategic} learning
that is  practical and effective,
and which we analyze theoretically.
We conclude with a set of experiments
that empirically demonstrate the utility of our approach.

\end{abstract}

\section{Introduction}
\label{sec:intro}

Machine learning is increasingly being used in domains where
human users are the subject of prediction.
But when users stand to gain from certain predictive outcomes,
they may be prone to act in ways that promote the outcomes they desire.
A growing recognition of this idea has led to much recent interest
in methods that are able to account for how humans respond to learned models.
One such line of research considers the problem of \emph{strategic classification} \citep{BrucknerS11,hardt2016strategic},
in which users can modify their features (at some cost) to obtain favorable predictions,
and the goal in learning is to be robust to such behavior.
Common tasks include loan approval, university admissions, and hiring---all examples in which users have incentive to be classified positively,
and in which systems must anticipate user behavior in order to predict well.
Strategic classification is appealing as a learning problem
in that it is simple yet succinctly captures a natural form of tension that can arise between systems and their users.
This has made it the focus of many recent works
\citep{sundaram2021pac,zhang2021incentive,levanon2021strategic,ghalme2021strategic,jagadeesan2021alternative,zrnic2021leads,estornell2021unfairness,lechner2021learning}.

But despite the elegant way in which it extends standard binary classification,
strategic classification 
remains narrow in the scope of strategic behavior it permits.
A key working assumption is that while 
the system is interested in correct predictions,
users are interested in \emph{positive} predictions.
One artifact of this is that learning is 
essentially restricted to 
settings in which one outcome is globally ``good'' for users (e.g., loan approved), while the other is ``bad'' (e.g., loan denied);
another is that 
users are grimly framed as always acting to ``game'' the system.
But not all outcomes are ``good'' or ``bad'',
and not all strategic behavior is gaming.
Here we argue for a broader perspective on
what constitutes strategic behavior in classification.

Towards this goal,
we propose and study the novel problem of
\emph{generalized strategic classification} (GSC),
which includes standard strategic classification (SC)
as a special case, as well as other rich problem sub-classes. 
Our formulation relies on the simple observation that
strategic behavior depends on three key elements:
(i) what users \emph{know},
(ii) what users \emph{want}, and
(iii) how users \emph{use} their knowledge to promote their goals.
For example, in standard strategic classification,
users know their true features (whereas the system does not);
want positive predictions (rather than correct predictions);
and are willing to invest effort or resources to obtain them
(by applying costly feature modifications).
But other forms of knowledge, aims, and means
lead to other types of strategic interactions between a system and its users.
The power of our framework lies in providing a simple handle for
reasoning about the effects of general forms of interaction on learning.

Within the space of GSC problems,
of particular interest to us is a sub-class of problems we refer to as \emph{incentive-aligned strategic classification} (IASC).
Aligned incentives naturally arise in settings
where predictions are intended to assist users;
in particular, they are \emph{for} users,
rather than \emph{about} users
(as in loans, hiring, etc.).
%
Applications in which incentives align are widespread---recommendation systems, search engines, and online marketplaces are all examples in which
prediction is provided by the system as a service to its users,
and both parties are interested in accurate predictions. 

From a modeling perspective, the key difference between SC and IASC
is simply that instead of wanting positive predictions,
users now want correct predictions---just as the system does.
But from a learning perspective, this mild change
transforms the learning problem from one of robustness (to gaming)
to one of cooperation,
and the key challenge in learning now lies in how to best utilize
individual strategic behavior to promote collective improvement.
Note this has concrete implications for transparency:
to successfully promote its own goals,
it should be in the best interest of the system to clearly communicate its predictive model to users, so that strategic updates are maximally effective
(c.f. SC, where the implications of transparency can be ambiguous \citep{ghalme2021strategic}).

How should learning be done in incentive-aligned settings?
A natural approach would be to simply take the conventional ERM rule
and replace the original features, $x$,
with their strategically-modified counterparts, $x'$.
Indeed, this approach is sensible when applied to the \emph{true error},
i.e., the 0/1-loss.
However, our first key result shows that the argument breaks
once this \naive\ approach is applied to a \emph{proxy loss},
such as the hinge loss.
This has concrete practical implications; as we show,
this \naive\ approach is prone to adverse generalization issues,
and is unnecessarily computationally complex.
As a solution, and with initial focus on a special instance of IASC,
we propose a new \emph{strategic hinge loss},
which builds on an adaptation of the notion of margin to strategic settings.
The strategic hinge has a simple and differentiable form,
and so can be optimized end-to-end.
For generalization, we give Rademacher-based bounds showing
that the strategic hinge loss enjoys favorable guarantees.



Building on these results,
we return to the most general class of GSC problems,
and show that the principles underlying our analysis for IASC
hold more broadly;
i.e., that naively applying the hinge loss to strategic inputs is not only suboptimal, but can nullify the concept of `margins' entirely.
We extend both of our modeling and theoretical contributions to the general case.
First,
we give a formula for constructing an appropriate strategic hinge proxy
for any instance of GSC,
this relying on a careful reinterpretation of the definition of margin.
We show that, like for IASC, the strategic proxies
for several notable sub-classes are simple and differentiable,
and have intuitive interpretations.
Second,
we extend our generalization bounds to hold for GSC in its broadest form;
we also give specialized bounds for several sub-classes,
and study their relations.
Our bounds have a tight connection to, and closely match,
standard non-strategic bounds.




We conclude with an empirical section that 
includes two sets of experiments.
The first set of experiments target generalization,
and in this way, complement our theoretical findings.
Here we evaluate performance on synthetic data and for
several types of strategic environments.
The second set of experiments consider an elaborate incentive-aligned
environment, designed to mimic a recommendation setting
in which users act on the basis of private, personalized past experiences.
For this setting, we use data based on a real fashion-related dataset (with simulated user responses).
The strategic hinge in this setting does \emph{not} have a simple form;
nonetheless, we propose a tractable approach for its optimization.
Our results demonstrate that our approach is efficient and effective.
Code is publically available at \url{https://github.com/SagiLevanon1/GSC}.

Taken together, our practical and theoretical results
suggest that GSC is an appropriate, well-balanced generalization:
it is flexible in the strategic behavior it permits
and the learning problems that follow,
but at the same time,
is precise enough to preserve general statistical
properties and empirical phenomena of conventional classification tasks.

\subsection{Related work}
\label{sec:related}



The literature on strategic classification is growing rapidly.
Various formulations of the problem were studied in earlier works
\citep{bruckner2009nash,bruckner2012static,grosshans2013bayesian},
but most recent works adopt the core setup of \citet{hardt2016strategic}.
Some studies focus on theoretical aspects;
for example, \citet{zhang2021incentive} and \citet{sundaram2021pac}
extend VC theory to account for strategic behavior.
Other works focus on practical aspects,
such as \citet{levanon2021strategic} who propose
a differentiable learning framework,
and \citet{eilat2022strategic} who study strategically-robust learning for graph-based classifiers.
Our work includes a blend of both theory and practice.

There have also been efforts to extend strategic classification beyond its original formulation.
For example,
\citet{ghalme2021strategic} study a setting where users respond based on
individually-estimated classifiers,
and \citet{jagadeesan2021alternative} consider users having noisy estimates of model parameters.
\citet{sundaram2021pac} make the connection to adversarial learning,
and propose a unified model allowing for more general forms of user response.
Our framework subsumes these earlier settings,
and generalizes beyond.

Several works in the adversarial literature have suggested
margin maximization as a means to achieve robustness
\citep{ding2020mma,yan2012adversarial}.
Our work extends beyond this idea and studies the general role
margins play under strategic behavior,
theoretically and methodologically.

Finally, we note there are many works within the broader area
of strategic-aware learning, focusing on
dynamics and online learning \citep{DongRSWW18,CLP19,zrnic2021leads},
causal effects \citep{miller2020strategic,shavit2020causal,lookahead,harris2021strategic,bechavod2021gaming},
strategic regression \citep{tang2021linear,harris2021stateful,bechavod2021information},
strategic ranking \citep{liu2021strategic},
and strategic representation \citep{nair2022strategic}.
These remain outside the scope of our generalized framework for
strategic classification.

\section{Generalized Strategic Classification}
\label{sec:gsc}


We begin by reviewing the learning setup for SC,
and then proceed to present GSC and IASC.

\subsection{Strategic classification}
Denote by $x \in \X \subseteq \R^d$ features representing user attributes,
and by $y \in \Y = \{-1,1\}$ their corresponding labels.
Let $D$ be some unknown joint distribution over
$\X \times \Y$.
As in standard classification,
at \emph{train time}
the system is given access to a sample set of $m$ labeled pairs $S=\{(x_i,y_i)\}_{i=1}^m$
sampled iid from $D$,
on which it aims to learn a classifier $h : \X \rightarrow \Y$ from some function class $H$.
What differs in strategic classification is that
at \emph{test time},
once $h$ has been learned $h$ (and deployed),
users can \emph{modify} their features
$x \mapsto x'$,
in a way we explain shortly.
The objective of the system is to
correctly classify \emph{modified} inputs,
and so its goal in learning is to minimize
the expected 0/1 loss,
$\err(h)=\expect{D}{\1{y \neq h(x')}}$.
Learning is therefore successful if it
produces a classifier that is robust to strategic modifications.


The way in which users modify their features
derives from the utility they can gain from predictive outcomes.
In SC, users are willing to modify features if this will help them be classified positively, 
denoted $\yhat=1$.
But modifications are costly,
and the cost of applying $x \mapsto x'$ is given by a \emph{cost function}
$c(x,x')$, which is assumed to be known to all.
Users are modeled as rational utility-maximizers that
update their features via the following \emph{best-response mapping}:
\begin{equation}
\label{eq:response_mapping_sc}
    \decsc_{h}(x) \triangleq \argmax_{x' \in \X} h(x') - c(x,x')
\end{equation}
Learning then aims to
minimize the expected \emph{strategic error}:
\begin{equation} \label{eq:sc_objective}
\argmin_{h \in H} \expect{D}{\1{h(\decsc_{h}(x)) \neq y}}
\end{equation}

\subsection{Generalizing strategic classification}
The update in Eq. \eqref{eq:response_mapping_sc}
includes three important ingredients: what users know (i.e., $x$), what they want (i.e., $h(x')=1$), and how these two concepts are integrated into action
(i.e., maximizing utility minus cost).
But the update is \emph{effective} 
only when the original features $x$ remain \emph{private user information}.\footnote{Indeed, had the system known $x$ at test time, any modification would have been futile (and costly).}
Privacy of information is a crucial aspect of strategic classification,
and is essentially what makes it distinct from standard classification.
Building on this simple observation, we propose a generalized framework
permitting users to make use of additional and richer forms of private information,
in more varied ways, and towards broader goals.

Concretely, our GSC framework extends SC in two ways.
First, we allow users to hold, in addition to $x$,
private \emph{side information}, denoted $z \in \Z$
(and redefine $D$ to be a joint distribution over triples $(x,z,y)$).
Objects $z$ can be of any type,
and the main distinction between $z$-objects and $x$-objects
(which are also private) is that while $x$ are used by the system as input to the classifier (and so are subject to modification by users),
$z$ are used only by the user, and only for the purpose of computing $\dec$
(and so remain unmodified).
One interpretation of $z$ is that it defines a user's `type'
(in the game-theroretic sense), which affects her actions.

Second, we allow for broader user objectives
by replacing $h(x')$ in Eq. \eqref{eq:response_mapping_sc}
with a more general function of \emph{perceived utility},
denoted $\putil_h(x',z)$,
which encodes the utility users \emph{believe} they obtain 
from $x'$ under the classifier $h$, and as a function of $z$.
This gives the \emph{generalized response mapping}:
\begin{equation}
\label{eq:response_mapping_gsc}
    \decgsc_{h}(x) \triangleq \argmax_{x' \in \X} \putil_h(x',z) - c(x,x')
\end{equation}
To see how $u$ and $z$ work together, observe that 
Eq. \eqref{eq:response_mapping_gsc} provides a formula for how users \emph{act},
meaning that $u$ reflects how utility is \emph{perceived}
by users.
This perceived utility may, or may not, align with users' \emph{true} utility (which must be independently defined);\footnote{In SC, they are the same.}
note this means $\decgsc$ is no longer necessarily a best-response.
What controls the degree to which true and perceived utility align is how useful
$z$ is as side information when integrated into $u$ to form user beliefs regarding value.
Side information can be helpful---but also misleading;
a simple example is when $z \in \R$ and acts as additive noise,
i.e., $\putil_h(x',z)=\util_h(x')+z$ for true utility $\util_h$,
as in Random Utility Theory \citep{Cascetta2009}.
But Eq. \eqref{eq:response_mapping_gsc} is more general,
and can in principle allow for more elaborate forms of user modeling, 
such as bounded-rational, Bayesian, or behavioral decision models.
We next survey some special instances of Eq. \eqref{eq:response_mapping_gsc}
which exemplify this idea.



\paragraph{Notable special cases.}
Standard SC (Eq. \eqref{eq:response_mapping_sc}) is obtained from Eq. \eqref{eq:response_mapping_gsc} by
setting $z=\emptyset$ (i.e., no additional side information) and $u_h(x',z)=h(x')$.
Notice Eq. \eqref{eq:response_mapping_sc} can be rewritten as:
\begin{equation}
\label{eq:response_mapping_sc_v2}
    \dec^{\mathtt{SC}}_{h}(x) \triangleq \argmax_{x' \in \X} \1{h(x')=1} - \frac{1}{2} c(x,x')
\end{equation}
where now $\putil_h(x,z)=\1{h(x')=1}=\frac{h(x')+1}{2}$.
This simply emphasizes that users want positive predictions,
but makes it easy to consider other forms of user interests.
For example, for \emph{adversarial} users,
we plug in $y$ and obtain:
\begin{equation}
\label{eq:response_mapping_adv}
    \dec^{\mathtt{adv}}_{h}(x,y) \triangleq \argmax_{x' \in \X} \1{h(x') \neq y} - \frac{1}{2} c(x,x')
\end{equation}
which is made possible once we set $z=y$ as private information.
More generally, if we let $z \in \{-1, 1\}$ hold arbitrary values,
then we recover the general-preference (GP) model studied in \cite{sundaram2021pac}:
\begin{equation}
\label{eq:response_mapping_gp}
    \decgp_{h}(x,z) \triangleq \argmax_{x' \in \X} \1{h(x') = z} - \frac{1}{2} c(x,x')
\end{equation}
here, knowing $z$ is essential for users to effectively respond.

Returning to SC, the recent work of \cite{jagadeesan2021alternative}
on alternative microfoundations
proposes a noisy response model.
This can be recovered by setting
$z \in \R^d$ as the random noise term, which gives:
\begin{equation}
\label{eq:response_mapping_mf}
    \dec^{\mathtt{noise}}_{h}(x,z) \triangleq \argmax_{x' \in \X} h_{\theta+z}(x') - c(x,x')
\end{equation}
where $\theta \in \R^d$ are the parameters of the classifier.
A more elaborate form of side information can be found in 
\cite{ghalme2021strategic},
where users are `in the dark' and
respond based on an approximate classifier, $\tilde{h}$,
estimated from data:
\begin{equation}
\label{eq:response_mapping_dark}
    \dec^{\mathtt{dark}}_{h}(x,z) \triangleq \argmax_{x' \in \X} \tilde{h}(x';z,h) - c(x,x')
\end{equation}
Here, $\tilde{h}(\cdotp;z,h)$ is estimated by users from a
sample set $\{(x_j,\yhat_j)\}_{j=1}^n$
consisting of additional examples $z=\{x_j\}_{j=1}^n$ labeled by the true classifier, $\yhat_j=h(x_j)$.
In both of the above, users seek positive predictions,
but act based on how they \emph{perceive} utility,
as determined by $u$ and $z$.

\subsection{Learning in GSC}
In its most general form, the GSC learning objective is:
\begin{equation} \label{eq:gsc_objective}
\argmin_{h \in H} \expect{D}{\1{h(\decgsc_{h}(x,z)) \neq y}}
\end{equation}
where for each sub-class of problems, $\decgsc$ is replaced
with its appropriate counterpart.
One key difference between Eq. \eqref{eq:gsc_objective}
and Eq. \eqref{eq:sc_objective} is that the former now
includes side information $z$.
This requires us to be precise about its role.
Clearly, at test time, neither $x$ nor $z$ are observed by the system.
However, at train time, it remains a question whether $z$ should be observed or not.
For example, in the GP setting of \cite{sundaram2021pac},
$z$ is assumed to be \emph{known};
in the noisy-parameter setting of \cite{jagadeesan2021alternative},
when $z$ models users' misperceptions of $\theta$,
it is reasonable to consider it as \emph{unknown};
while for the `in-the-dark' setting of \cite{ghalme2021strategic},
it is shown that for known $z$, learning reduces to standard strategic classification,
whereas for unknown $z$, errors in estimating user responses can blow up
in a way that can cause learning to fail completely.
Hence, the decision regarding observing $z$ remains task-specific.

In this paper, we focus on the setting where $z$ is known---this places emphasis
on the introduction of $z$ into the response mapping $\dec$,
and abstracts away other aspects (such as what system knows or does not).
Note also that when $z$ is unknown, from the system's perspective,
the user's response mapping $\dec$ (Eq. \eqref{eq:response_mapping_gsc})
is not well-defined, and so requires additional assumptions on how 
the system compensates for this lack of knowledge in learning,
which we aim to avoid.
We can therefore think of $z$ as obtained at train time in the same manner that ground-truth labels $y$ (which are also unobserved at test time) are typically gathered, e.g., by investing effort to collect them from users.





\section{Incentive-Aligned Strategic Classification} \label{sec:iasc}
Next, we introduce and study a particular sub-class of problems
in GSC which we refer to as \emph{incentive-aligned strategic classification} (IASC).
Our results here will serve as a basis for our
more general results for GSC in Sec. \ref{sec:theory},
but are of independent interest (full proofs in the Appendix).

Intuitively, incentive alignment occurs when the utility of users is similar enough to that of the system so that strategic behavior improves both.
This can happen when users also seek correct predictions,
and the system provides such predictions as a service
(recommendation systems are one such example).
When incentives align, strategic behavior can improve outcomes for both the system and its users.

We begin by formally defining \emph{incentive alignment} (IA).
\begin{definition}[Incentive-alignment] \label{def:ia}
Let $D$ be a joint distribution over $\X\times\Z\times\Y$,
$H$ be a function class, and $\dec$ a response mapping.
We say the learning task is \emph{incentive-aligned} if
$\exists \, h \in H$ such that $\forall \, h' \in H$, it holds that:
\begin{equation} \label{eq:ia_def}
\mathbb{E}_{D}[\1{h(\dec_h(x;z)) \neq y}] \leq
\mathbb{E}_{D}[\1{h'(x) \neq y}] 
\end{equation}
If such an $h$ exists, we say it \emph{aligns incentives}.
\end{definition}
Incentive alignment holds when there is some classifier 
whose performance on strategically modified inputs 
is better than the optimal classifier on unmodified inputs. 
%
We next describe two novel settings
in which users seek to accurate predictions,
and so incentive alignment is likely.

\textbf{Noisy labels (NL).}
In this simple model, side information $z$ consists of a noisy estimate of a user's true label, denoted $\ytilde \in \{\pm 1\}$.
Users seek accurate predictions (i.e., want $\yhat = y$),
but act as if $\ytilde$ (which may be $\neq y$) is their true label:
\begin{equation}
\label{eq:response_mapping_nl}
\decnl_h(x, \ytilde) = \argmax_{x' \in \X} \1{h(x')=\ytilde} - \frac{1}{2}c(x,x')
\end{equation}
When $\ytilde$ and $y$ are sufficiently correlated,
and if the system is able to 
`correct' erroneous user updates,
then incentive alignment should hold.
Note $\decnl$ is a special case of $\decgp$.


\textbf{Personalized previous experiences (PPE).}
Here we mimic a recommendation setting
in which the system provides users $x$ with personalized (binary) relevance predictions $y \in \{\pm 1 \}$ for various items $a \in \R^\ell$.
Instances are tuples $(x,z,a,y)$ ($D$ is extended accordingly),
and the system aims to predict relevance for user-item pairs, $\yhat=h(x',a)$.
For modifying features, users have as side information
previously experienced items,
in the form of a sample set $z=\{(a_j,y_j)\}_{j=1}^n$, where typically $n \ll m$.
Here,  $\decppe_h  (x,z)$ is:
\begin{align}
\label{eq:response_mapping_ppe}
\argmax_{x' \in \X} \frac{1}{n}\sum\nolimits_{j=1}^n\1{h(x', a_j) = y_j} - \frac{1}{2}c(x,x') 
\end{align}
Hence, users modify their features to maximize the accuracy of the proposed $h$
on their past preferences.

In the settings above, the system and its users cooperate (indirectly)
to achieve better accuracy,
and are complementary in their strength:
the system has large data but on the collective of past users,
whereas new users have small or noisy data,
but personalized for their own preferences.
Learning is successful if the system can harness user efforts to improve
accuracy for all.
%
%
In the remainder of this section we focus on the noisy labels model
to establish our main results for IASC,
which we then extend to generic GSC cases in Sec. \ref{sec:theory}.
We return to PPE (and define it more precisely) in Sec. \ref{sec:experiments}.

\subsection{Learning with noisy labels as side information}
\label{sec:learning_nl}

\begin{figure}[t!]
\includegraphics[width=\columnwidth]{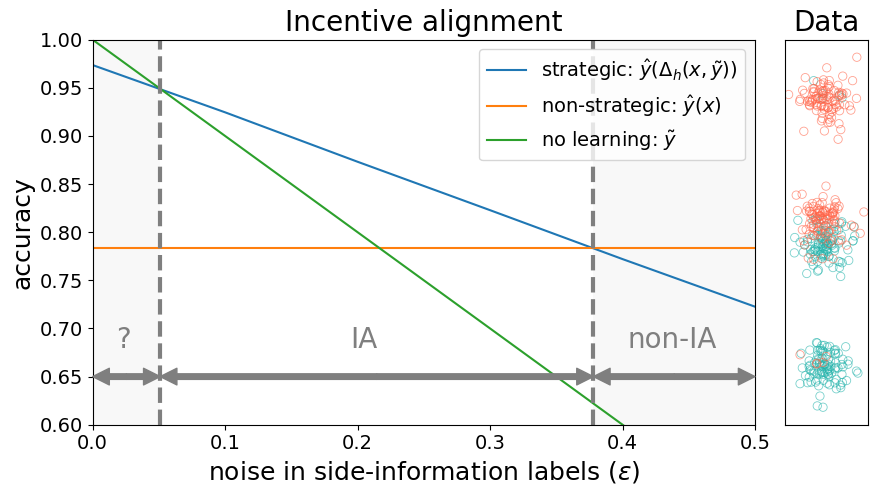}
\caption{Regions of incentive alignment in NL,
for synthetic data (see right plot) and varying noise $\epsilon$.
For a wide range of $\epsilon$, learning is incentive-aligned (IA),
meaning that the strategic behavior of users is beneficial for both
the system and its users.
The leftmost region is also IA, but $\ytilde$ is better for users than $\yhat$.
}
\label{fig:varying_epsilon}
\end{figure}

In this section we study an instance of the noisy-label side-information
setting in which noise is uniform and independent:
$\ytilde=y$ w.p. $1-\epsilon$ and $\ytilde=1-y$ w.p. $\epsilon$ for all $y$.
The noise parameter $\epsilon$ controls how informative $\ytilde$ are of $y$;
this has direct connections to incentive alignment.
Figure \ref{fig:varying_epsilon} shows on synthetic data
that incentives align for most values of $\epsilon$ (Def. \ref{def:ia}),
and hence strategic behavior is preferable
(details in Appendix \ref{sub:varying_exp}).
In Appendix \ref{sec:ia_conditions},
we theoretically characterize the conditions under which incentive alignment holds,
and under which predictions are informatively useful.

Throught this section we focus on linear classifiers
$h_{w}(x)=\sign(f_{w}(x))$ where $f_{w}(x)= w^\top x + b$,
and on 2-norm costs $c(x, x') = \norm{x-x'}_2$.
We use $\dec_h$ to mean $\dec_{h_w}$,
and omit the intercept term $b$ throughout for clarity.

\paragraph{Learning via loss minimization.}
A natural approach to minimizing the expected strategic loss (Eq. \eqref{eq:gsc_objective} with $\decnl$)
is to instead aim at minimizing the empirical risk:
\begin{equation} \label{eq:nl_empirical_objective}
\argmin_{w \in \R^d} \frac{1}{m}\sum_{i=1}^m \1{h_w(\decnl_{h}(x_i,z_i)) \neq y_i}
\end{equation}    
While this approach is well-motivated \emph{in principle},
in practice, since the 0/1 loss is intractable;
just as for non-strategic data,
a proxy loss is needed, and to control for overfitting,
regularization is added.
Using the hinge loss as a proxy and $L_2$ regularization (as in SVM),
{\naive}ly adapting the standard objective to the strategic case gives:
\begin{equation}
\label{eq:objective_emp}
\argmin_{w \in \R^d} \frac{1}{m} \sum_{i=1}^m \max\{0, 1- y_i w^\top \decnl_{h}(x_i,z_i))\} + \lambda \norm{w}_2^2
\end{equation}

Eq. \eqref{eq:objective_emp} seems appealing and straightforward:
in the same way as for the 0/1 loss,
it simply replaces $x$ with $\decnl_h(x,z)$.
However, as we show in the next section, this \naive\ approach
has two significant drawbacks:
it is difficult to optimize,
and may fail to generalize well.
Hence, while minimizing the strategic empirical 0/1 loss is a sound approach,
a \naive\ adaptation of conventional proxy approaches may not be.


\paragraph{The problem with margins.}
Consider first a non-strategic setting.
If data is linearly separable, then
multiple classifiers can exist having zero empirical error.
But some solutions are better than others,
and so to guarantee good generalization, 
additional criteria (beyond loss minimization) must be considered.
One popular approach is \emph{max-margin learning} \citep{cortes1995support},
which aims to learn the classifier having the largest \emph{margin}:
\begin{equation}
\label{eq:margin}
\margin_w \triangleq \min_{i \in [m]}  y_i \wbar^\top x_i, \quad \qquad 
\wbar = w / \norm{w}
\frac{w}{\norm{w}}
\end{equation}
The main motivation for max-margin learnin is that large-margin classifiers
can be shown to generalize well,
and the canonical max-margin approach (for linearly-separable data) is Hard-SVM. 
A \naive\ adaptation of Hard-SVM to our strategic setting
would be to replace each $x_i$ with $\dec_h(x_i,\ytilde_i)$, this giving:
\begin{align}
\label{eq:naive_ssvm}
 \argmax_{w\,:\,\norm{w} = 1} \, & \min_{i \in [m]} 
\left| w^\top \dec_h(x_{i}, \ytilde_{i})  \right| \nonumber \\
& \text{ s.t.}  \,\,\, y_{i} w^\top \dec_h(x_{i}, \ytilde_{i}) > 0 \quad \forall i \in [m] 
\end{align}

\begin{figure}
\includegraphics[width=\linewidth]{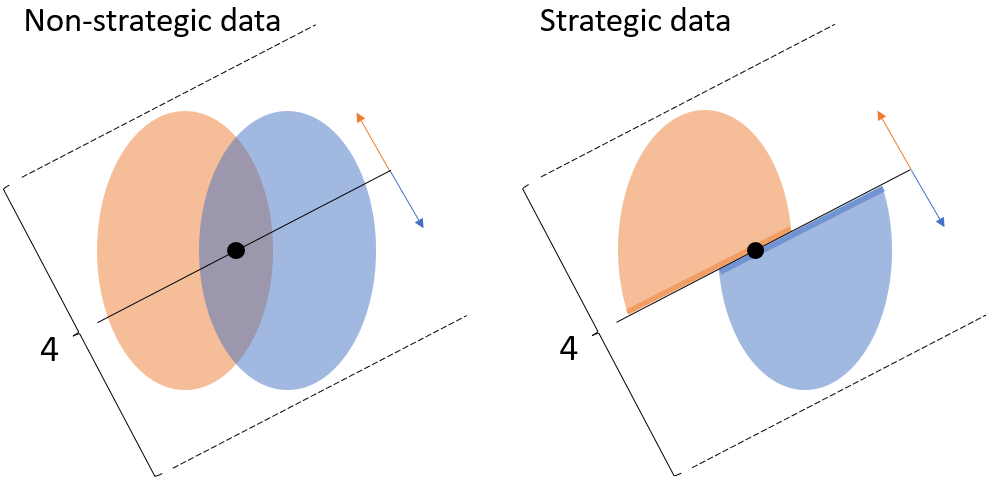} 
\caption{
\textbf{(Left)}
In non-strategic settings, maximzing the (soft) margin is a useful criterion for learning models that generalize well.
\textbf{(Right)}
In strategic settings, this logic breaks;
even for strategically separable data,
once any point moves, the margin becomes zero.
One implication is that mulitple models are likely to attain the max margin (which is zero).
The illustration portrays one such model,
but \emph{all} models (passing at the origin) are equally `good'.
}
\label{fig:multiple_sol}
\end{figure}

Note that since points now `move',
data that was originally separable may no
longer satisfy the constraints in Eq. \eqref{eq:naive_ssvm}.
Conversely, non-separable data might \emph{become} separable
if points move in a way that `corrects' classification mistakes
(e.g., consider the extreme case of $\ytilde=y$).
This calls for
a notion of \emph{strategic} separability.
\begin{definition} 
\label{def:strategic_separability}
A sample set 
is \emph{strategically linearly separable} if there exists a 
$w$ that separates the data \emph{after} it has been modified, i.e.,
$y_{i} w^\top \dec_h(x_{i}; z_{i}) > 0$
$\forall i \in [m]$.
\end{definition}
Intuitively, a model $w$ that (strategically) separates the data and has a large margin
should be appealing.
However, a key observation is that for any $w$ that permits \emph{some} movement---the maximal margin is \emph{zero};
this is since users minimize costs, and so land directly on the decision boundary.
This has concrete implications on learning.
Consider, for example, data that is strategically linearly separable,
but is not (non-strategically) linearly separable\footnote{This is sufficient; strategic linear separability is necessary.}.
In this case,
\emph{multiple optimal solutions are likely to exist}---all of which have a margin of zero---and to which the learning algorithm is oblivious (see Figure \ref{fig:multiple_sol}). 
Thus, the \naive\ 
approach resurfaces the very same issues
that the max-margin approach aims to \emph{avoid}.
This deficiency is easiest to imagine for $\ytilde=y$,
but holds for GSC settings beyond NL (Sec. \ref{sec:experiments_generalization}).

\paragraph{Strategic margins and hinge.}
Our next result shows that to recover the generalization properties of the max-margin approach,
we must rethink the notion of `margin'.
We begin by reinterpreting the strategic constraints in Eq. \eqref{eq:naive_ssvm}.
\begin{lemma}
\label{lem:s-margin}
For any linear classifier $w$, it holds that:
\begin{equation*}
y_i w^\top \decnl_h(x_i,\ytilde_i) >0
\,\,\Leftrightarrow \,\,
y_i(w^\top x_i + 2\ytilde_i\|w\|) >0
\end{equation*}
\end{lemma}
Proof in Appendix \ref{pr:s-margin}.
The constraints can therefore be rewritten in a simple form that does not 
rely on $\dec$.
This suggests that margins should be measured accordingly.
\begin{definition} 
\label{def:s-margin}
The \emph{strategic margin} (or \emph{s-margin}) of $w$ is:
\begin{equation}
\label{eq:s-margin}
\tmplt{\margin}{\mathnormal{w}}{NL} \triangleq \min_{i \in [m]} y_i( \wbar^\top x_{i} + 2 \ytilde_i),
\quad \qquad \wbar = 
w / \norm{w}
\end{equation}
\end{definition}

\begin{figure}[t!]
\centering
\includegraphics[width=0.7\columnwidth]{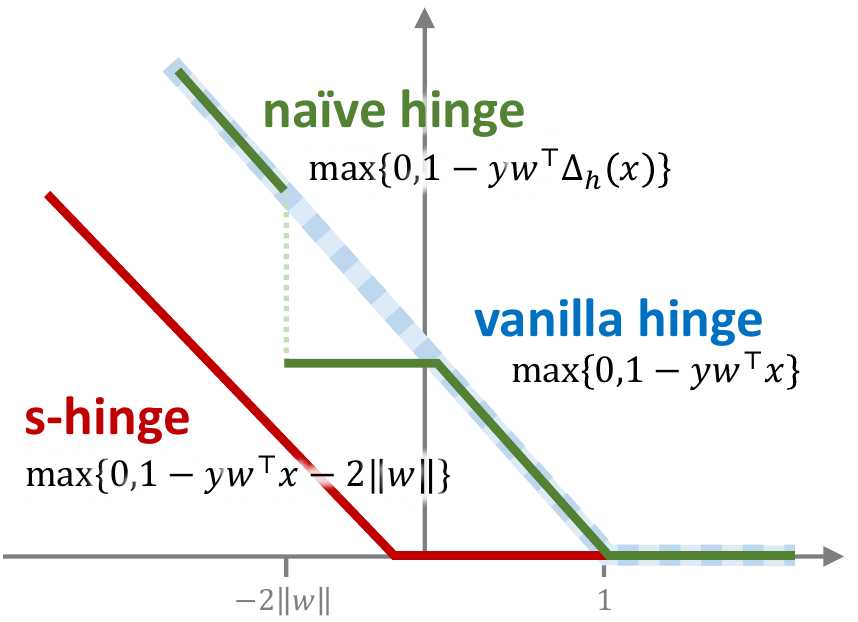}
\caption{An illustration of the s-hinge for NL with $\ytilde=y$.
Strategic movement creates a flat `kink' in \naive\ hinge
at $[-2\norm{w},0]$, since all points in this range move
to the decision boundary.
The s-hinge corrects for this by properly anticipating strategic behavior. 
}
\label{fig:hinges}
\end{figure}

Note that Eq. \eqref{eq:s-margin} differs from
the standard margin 
in the corrective term $+2\ytilde$,
which captures the maximal admissible cost, and the direction of change.
This new margin has an intuitive interpretation:
the distance of each $x_i$ is measured w.r.t.
an `individualized` classifier, $w^{(i)}$,
which is simply $w$ shifted parallelly by 2 units in the direction of anticipated movement, $\ytilde_i$ (see Fig. \ref{fig:s-margin} in Appx. \ref{apdx:illustrations}).
This new margin is \emph{not} equivalent to 
the \naive\ objective in Eq. \eqref{eq:naive_ssvm}.

For strategically-separable data,
our approach will be to optimize the s-margin using the appropriate constraints.
For non-separable problems,
this can easily be extended to the `soft' case
using slack variables, akin to standard soft SVM (see Appendix \ref{sub:soft_ssvm}).
From this, we can derive an appropriate proxy loss---the \emph{strategic hinge}, or \emph{s-hinge}.
\begin{definition} \label{def:s-hinge}
For noisy labels,
the \emph{strategic hinge} is:
\begin{equation*}
\tmplt{\loss}{s-hinge}{NL}(x,\ytilde,y;w) \triangleq
\max\{ 0, 1-y(w^\top x + 2 \ytilde \norm{w}) \}
\end{equation*}
\end{definition}
Hence, the s-hinge is simply the standard `vanilla' hinge, but with an additional additive term of $-2y\ytilde\norm{w}$,
which can be interpreted as aiming to correctly classify \emph{unmodified} inputs $x_i$,
but w.r.t. an instance-dependent reference point, $1-2\ytilde\|w\|$,
from which the margin is measured.
To see where this comes from, 
note that 
the s-hinge can be rewritten as:
\begin{equation*}
\label{eq:s-hinge_alt}
\max \{ 0, 1 - yw^\top \decnl_h(x,\ytilde) - 
(2-c(x,\decnl_h(x,\ytilde)))y\ytilde\norm{w}\}
\end{equation*}
(proof in Appendix \ref{pr:s-hinge}).
This is exactly the \emph{\naive} hinge,
$\max \{ 0, 1 - yw^\top \decnl_h(x,\ytilde)\}$,
but with a term that accounts for
both the \emph{actual} and \emph{maximal} cost. 
Figure \ref{fig:hinges} illustrates the crux of the \naive\ hinge,
and how the s-hinge corrects it.

Finally, adding regularization gives our learning objective:
\begin{equation}
\label{eq:ss-svm}
\argmin_{w \in \R^d} \frac{1}{m} \sum_{i=1}^m
\tmplt{\loss}{s-hinge}{NL}(x_i,\ytilde_i,y_i;w)
+\lambda \norm{w}_2^2
\end{equation}
Our approach has two key benefits:
it can be efficiently optimized,
and it enjoys favorable generalization guarantees (Sec. \ref{sec:gsc_bounds});
fortunately, and as we show next,
our approach (and its benefits) extend to broader instances of GSC.

\begin{figure*}[t!]
\centering
\includegraphics[width=\textwidth]{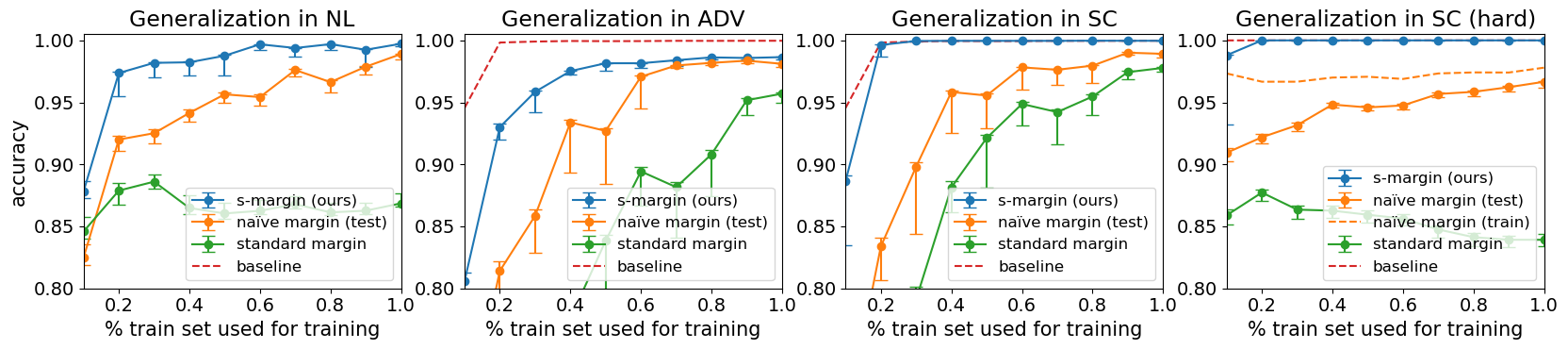}
\caption{Results for the generalization experiment.
The three leftmost plots are on `easy' environments 
that reflect differences in generalization behavior.
The rightmost plot is on a `hard' environment
in which the \naive\ approach also suffers from optimization issues.
}
\label{fig:generalization}
\end{figure*}

\section{GSC: Proxy Losses and Bounds} \label{sec:theory}


In this section, we generalize our approach and results to GSC,
this using the principles underlying our results for NL in Sec. \ref{sec:iasc}.
Our focus remains on linear classifiers, but we will now allow for general side information $z$, utilities $\putil$, and costs $c$.
In particular, our results here cover the personalized previous experiences model (PPE) proposed in Sec. \ref{sec:iasc}. 

\subsection{Generalized s-hinge}
Returning to the standard hinge,
note it can be rewritten as:
\begin{equation}
\max\{0, 1 - \sign(yw^\top x)|w^\top x| \}
\end{equation}
which decouples the original penalty term $yw^\top x$ into a `correctness' term,
$\sign(yw^\top x)=\1{y=h(x)}$,
and an un-normalized `distance' term,
\begin{equation}
\label{eq:distance}
d(x;w) \triangleq | w^\top x| = | \wbar^\top x | \norm{w}
\end{equation}
Now, consider again the \naive\ hinge. Under this perspective:
\begin{align}
\label{eq:naive_hinge}
& \max\{0,1-yw^\top \dec_h(x,z) \} \nonumber \\
= &\max\{0,1-\sign(yw^\top \dec_h(x,z))|w^\top \dec_h(x,z)|\}
\end{align}
This reveals its fault:
while it is appropriate to consider \emph{correctness} on modified inputs,
measuring \emph{distances} in this way lacks to account for the
fact that $\dec_h(x,z)$ is a \emph{response} to $w$;
in other words, it treats $\dec_h(x,z)$ as some predetermined fixed point $x'$,
discarding any information on the original point $x$ and how it changes as a function of $w$.




To appropriately adapt the hinge to strategic settings,
we must rethink the definition of margin.
Notice that the standard distance in Eq. \eqref{eq:distance}
can be reinterpreted as follows:

\textbf{Observation:}
\textit{$d(x;w)$ is the distance from $x$ to the closest point $x'$
that is classified differently, i.e.,}
\begin{equation}
\label{eq:distance_alt}
d(x;w) = \min_{x'} \|x-x'\| \,\, \text{s.t.} \,\, h(x)\neq h(x')
\end{equation}
This definition considers distances between \emph{points},
rather than between a point and a hyperplane, as in Eq. \eqref{eq:distance};
here, the role of the hyperplane is expressed in terms of classification outcomes
(as a constraint).

We can now adapt Eq. \eqref{eq:distance_alt} to a strategic setting
by replacing $x$ with $\dec_h(x,z)$ only where it affects predictions.
\begin{definition}
The \emph{strategic distance} of a point $x$ (and w.r.t. $z$)
from the decision boundary of $h_w$ is:
\begin{align}
\label{eq:s-distance}
d_\dec(x,z;w) = & \min_{x'} \norm{x-x'} \\
& \text{ s.t. } \,\, h(\dec_h(x,z)) \neq h(\dec_h(x',z)) \nonumber
\end{align}
\end{definition}
Here, unlike the \naive\ approach,
distance are correctly measured on \emph{un}-modified inputs,
and modified inputs are used to anticipate changes in classification outcomes.
These can then be used to define a \emph{generalized strategic margin}.

Consequently, strategic distances also suggest the appropriate form for
the \emph{generalized strategic hinge}, or \emph{gs-hinge}:
\begin{definition}
The \emph{generalized strategic hinge} is defined as:
\begin{align}
\label{eq:s-hinge_gsc}
& \tmplt{\loss}{s-hinge}{GSC}(x,z,y;w) \triangleq \\
& \quad\,\, \max\{0, 1 - y\sign(w^\top \dec_h(x,z))d_\dec(x,z;w)\norm{w} \} \nonumber
\end{align}
\end{definition}
which applies to any subclass of GSC.
Note that for the non-strategic setting,
plugging $\dec_h(x,z)=x$ into Eq. 
\eqref{eq:s-hinge_gsc} recovers the standard margin.
The rationale behind the gs-hinge for generic instances of GSC
is illustrated in
Fig. \ref{fig:s-loss} (see Appendix \ref{apdx:illustrations}).

\paragraph{Optimization.}
In its most general form, and with no particular structure,
optimizing the gs-hinge is likely to be hard.
However, for several subclasses of interest,
it turns out to have a tractable form.
For the special case of GP (which includes NL, SC, and ADV)
and for norm costs,
the strategic distance can be simplified to
$\tmplt{d}{\mathnormal{\dec}}{GP}(x,z;w)=
| \wbar^\top x + 2z |$
since points move at most distance 2
in the direction of $z$ (proof in Appendix \ref{pr:flipping_distance_gp}).
For this case, Eq. \eqref{eq:s-hinge_gsc}
becomes:
\begin{equation}
\label{eq:s-hinge_gp}
\tmplt{\loss}{s-hinge}{GP}(x,z,y;w) \triangleq 
\max\{0, 1 - yw^\top x -2zy \norm{w} \}
\end{equation}
For NL, $z=\ytilde$, which recovers $\tmplt{\loss}{s-hinge}{NL}$
(Def. \ref{def:s-hinge}).
For SC and ADV,
$z=1$ and $z=-y$, respectively.

For all cases above,
the gs-hinge in Eq. \eqref{eq:s-hinge_gp} accounts for strategic movement
\emph{but does not explicitly include $\dec_h(x,z)$ as a term}.
This removes the main computational difficulty which strategic behavior introduces---coping with an argmax.
In addition, Eq. \eqref{eq:s-hinge_gp}, is differentiable,
and so can be optimized end-to-end using standard gradient methods.
We view this as a substantial advantage of our approach.






\subsection{Generalization bounds} \label{sec:gsc_bounds}

Our next result provides a data-dependent bound on the
expected 0/1 error 
when minimizing the gs-hinge objective. 

\begin{theorem} 
\label{thm:bound_gsc}
Let $D$ be a joint distribution over triples $(x,z,y)$,
and let $r=\max_{x \in \X} \norm{x}$.
Denote by $\what$ the minimizer of the empirical generalized strategic hinge loss with $L_2$ regularization.
Then for every $\delta \in [0,1]$,
if the training set $\smplst$ includes $m$ samples,
then w.p. $\ge 1-\delta$ it holds that:
\begin{equation*}
\label{eq:bound_gsc}
    \tmplt{\Loss}{0/1}{} \leq 
    \tmplt{\Lossapx}{s-hinge}{GSC} +  \frac{8r\norm{\what}}{\sqrt{m}} 
    + (1+2\tmplt{\rho}{}{GSC} \norm{w}) \sqrt{\frac{2 \ln(\frac{4\norm{\what}}{\delta})}{m}}
\end{equation*}
where $\tmplt{\Loss}{0/1}{}$ is the expected 0/1 loss,
$\tmplt{\Lossapx}{s-hinge}{GSC}$ is the empirical s-hinge loss on $\smplst$,
and
$\tmplt{\rho}{}{GSC} = 2r$.
\end{theorem}
The proof (Appendix \ref{pr:bound_gsc}) relies on Rademacher bounds,
and carefully adapts the approach in \citep{BoundsProof}
to account for strategic updates.
The bound in Thm. \ref{thm:bound_gsc}
closely matches the original bound for non-strategic settings,
with the only differences being that in the original
bound the constant in the middle summand is 4 (vs. 8),
and $\rho=r$.
We view this as suggesting that, in some sense,
the gs-hinge is an appropriate strategic generalization
of the standard hinge.

For notable subclasses of GSC, the bound in Thm. \ref{thm:bound_gsc} can be tightened: the middle constant is reduced to the original 4,
and each case has its own $\rho$ term (Appendix \ref{pr:bounds}). 
For GP, we get $\tmplt{\rho}{}{GP}=
(r+2) \le \tmplt{\rho}{}{GSC}$,
and also $\tmplt{\rho}{}{SC}=\tmplt{\rho}{}{adv}=\tmplt{\rho}{}{GP}$
in the worst case.
For NL, we get $\tmplt{\rho}{\mathnormal{\epsilon}}{NL}=r-2+4\epsilon$;
hence, for $\epsilon=1/2$, 
the standard bound is recovered, and
for smaller $\underline{\epsilon} \leq \nicefrac{1}{2}$,
sample complexity in NL is \emph{better} since
$\tmplt{\rho}{\mathnormal{\underline{\epsilon}}}{NL} < \rho$;\footnote{Since points move within $\X$, and since movement is for distance of at most 2, we make the simplifying assumption that $r\ge2$.}\footnote{For larger $\epsilon \in (1/2,1]$, where most labels are flipped, sample complexity is worse.
Note $\epsilon=1$ matches an adversarial setting.}
and when incentives align,
the empirical loss is also likely to be lower.
Together, and for
$\overline{\epsilon} \ge \nicefrac{1}{2}$,
we get:
\[
\tmplt{\rho}{\mathnormal{\underline{\epsilon}}}{NL} \le
\rho \le \tmplt{\rho}{\mathnormal{\overline{\epsilon}}}{NL} \le
\tmplt{\rho}{}{SC} = \tmplt{\rho}{}{adv} = \tmplt{\rho}{}{GP}
\le \tmplt{\rho}{}{GSC}
\]






\section{Experiments} \label{sec:experiments}
We now turn to our experimental evaluation.
Our first experiment empirically studies generalization behavior
for the gs-hinge on synthetic data and in various strategic settings.
Our second experiment
returns to the personalized previous experiences (PPE)
setting from Sec. \ref{sec:iasc} and uses real data.

\subsection{Generalization} \label{sec:experiments_generalization}
To complement our theoretical results, 
we study the generalization behavior
of learning with the gs-hinge, as it compares
to the \naive\ hinge and standard (non-strategic) hinge,
across multiple settings: NL ($\epsilon=0$), SC, and adversarial (ADV).
We also add as a baseline the standard hinge applied to non-strategic data
(in SC and ADV it is an upper bound on strategic performance; in NL, it is a lower bound).
The challenge in empirically analyzing generalization is that
it is difficult to decouple statistical and optimizational issues;
this is particularly true in our case, 
since the gs-hinge (for the settings we consider) has both a simpler functional form
and better theoretical guarantees than the \naive\ hinge.
We therefore experiment in two types of synthetic environments:
`easy' and `hard', which vary in the difficulty to optimize.
Further details are in Appendix \ref{sub:generalization_exp}.

\paragraph{Results.}
Figure \ref{fig:generalization} presents our results.
Each plot shows performance for an increasing number of samples used for training.
Results are averaged over $30$ random splits,
with bars showing (asymmetric) standard errors. 
For the easy environment, the figure shows that across all settings,
both our gs-hinge and the \naive\ hinge achieve
roughly optimal accuracy when training on 100\% of the data;
however the gs-hinge strictly dominates the \naive\ hinge,
and converges significantly faster.
This result suggests that the gs-hinge generalizes better.
For the hard environment, results show that the
\naive\ approach does not reach the optimum;
train accuracy shows that this is not due to overfitting.
This shows that the simplicity of the gs-hinge is beneficial.


\subsection{Personalized Previous Experiences}
\label{sub:recsys}
In PPE,
the goal is to predict for a user-item pair 
$(x,a)$ whether user $x$ will like item $a$.
Users can modify their inputs
based on a (private) sample set of previous experiences,
$z=\{(a_j,y_j)\}_{j=1}^n$.
We use $c(x,x')=\norm{x-x'}_2^2$
and bi-linear classifiers $h_W(x, a) = \sign(a^\top W x)$,
where $w_x = Wx$ is a `personalized' classifier.

The PPE setting is unique in that, in essence, users
also aim to solve a classification problem.
To see this, note that for a given $W$,
user responses $\decppe_h (x,z)$
(Eq. \ref{eq:response_mapping_ppe})
becomes:
\[
\argmin_{x' \in \R^d} 
\frac{1}{n}\sum\nolimits_{j=1}^n\1{\sign(a_j^\top W x') \neq y_j} - \frac{1}{2} \norm{x-x'}_2^2
\]
This is a linear classification problem:
$x'$ act as learned classifier weights,
$a_j^{_{(W)}}=a_j^\top W \in \R^d$ serve as features,
$y_j$ are target labels,
and $x'$ is regularized towards $x$.
In this sense, PPE becomes a problem
of multiple agents solving interwoven classification problems,
and with similar goals.

\paragraph{Data.}
We use the Coats Shopping dataset from \citet{schnabel2016recommendations}.
The data includes 290 users and 300 items (coats),
with features for both (e.g., fashion preferences for users,
type and color for coats).
Labels include relevance scores (here, binarized)
given by each user to 40 coats.

\paragraph{Optimization.}
In principle, both system and users aim to solve difficult 0/1-loss
objectives; as these are in general intractable,
we model both entities as solving  proxy objectives instead.
For the system we use the gs-hinge in (Eq. \eqref{eq:s-hinge_gsc}).
In PPE, the general hinge is unlikely to have a closed-form solution.
This makes learning especially challenging, as it requires
solving a triple-nested optimization problem
(the outer $\argmax$ on $w$; the interim $\min$ in $d_\dec(x,z;w)$;
and the inner $\argmax$ in $\dec_h$).
Nonetheless, we show that if the system uses a squared-loss proxy,
the objective becomes tractable (for a full derivation see Appendix Sec. \ref{sub:ppe}).

\begin{figure}[t!]
\centering
\includegraphics[width=\columnwidth]{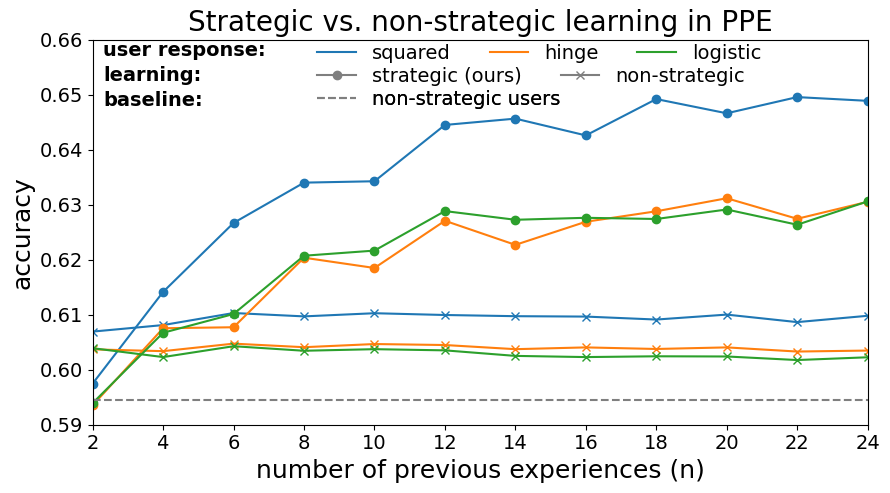}
\caption{
Results for the PPE setting on the coats dataset.
Learning with the strategic gs-hinge
is significantly better than the standard hinge,
starting at $n=4$ and improving as $n$ increases.
Performance is optimal when user respond with a proxy loss
that match the system's (squared).
Results show clear incentive alignment.
}
\label{fig:ppe}
\end{figure}

\paragraph{Results.}
We consider three user types that differ by the proxy loss they employ
to solve $\decppe$: square, hinge, and logistic.
For each user type, we compare the performance of our 
gs-hinge approach to the standard hinge,
and vary the number of previous experiences, 
$n=|z|$.
%
Figure \ref{fig:ppe} shows that for all user types,
and with as little as $n=4$ examples in $z$,
the gs-hinge provides a clear improvement over the standard hinge.
As $n$ grows, the gain in accuracy becomes more pronounced.
For the square-loss user, $n=24$ improves upon the standard hinge by $6.4$\%.
Improvement exhibits a clear diminishing-returns trend,
with most of the gain achieved by $n=12$ ($5.7$\%).
For the hinge- and log-loss users, gains are less pronounced,
but still significant ($3.8\%$ at $n=12$, after which gains plateau).
%
Note that both parties are better off when users employ the squared loss,
this matching the proxy loss used by the system.
Hence, it is in the best interest of the system to be transparent
about the loss it uses (here, squared), and in the best interest of users follow suit.
This is true since incentives align; results show clear incentive-alignment
from $n=4$. 
Interestingly, strategic behavior under incentive alignment also helps the standard (non-strategic) hinge classifier (vs. the baseline).




\section{Conclusions}  \label{sec:conclusions}

In a world where decisions about humans are increasingly being made by
(or with the support of) learned classifiers,
it is only reasonable to expect that humans will act to promote their own goals.
Our main takeaway is that these goals,
and the means taken to pursue them,
are varied.
Our generalized framework aims to capture this idea,
and to provide what we feel is a much needed flexibility
in user modeling within this domain.
Methodologically, our paper shows that coping with strategic behavior
cannot always be done by simply plugging a human response model
into exiting methods for non-strategic learning,
and that any adaptation of conventional approaches
must be done with care.

While our framework is general, it is intentionally restricted
to the fundamental setting of supervised binary classification.
Our hopes are twofold:
that within the limits of our framework,
its flexibility will aid in developing new and interesting 
strategic learning problems;
and that beyond our framework's boundaries,
these new problems can extend to broader settings
including dynamics, causality, and others.

\bibliography{refs}
\bibliographystyle{icml2022}

\newpage
\appendix
\onecolumn

\section{Additional Results}

\subsection{Conditions for Incentive Alignment in NL}
\label{sec:ia_conditions}

Our first results give a characterization of 
when a learning task is incentive aligned (see Definition \ref{def:ia})
in a simple instance of the noisy label model
with independent uniform noise:
$\ytilde=y$ w.p. $1-\epsilon$ and $\ytilde=1-y$ w.p. $\epsilon$ for all $y$.

As the noise parameter $\epsilon$ controls how informative $\ytilde$ are of $y$, we can ask: for what values of $\epsilon$ is strategic behavior
beneficial to all?
As we show, the answer directly depends on whether incentives can be aligned.

We begin with a definition of \emph{flipping cost}.
\begin{definition}
\label{def:flipping_cost}
The \textit{flipping cost} of a point $x$
with respect to a classifier $h$
and a cost function $c$ is:
\begin{equation}
\label{eq:flipping_cost}
\flipcost_{h}(x) \triangleq \min_{x' \in \X} c(x, x')
\quad \text{s.t.}\quad  h(x') \neq h(x)
\end{equation}
\end{definition}
Hence, $\flipcost_h(x)$ is the minimal cost required for modifying $x$
to `flip' its label (we will return to label flipping in Sec. \ref{sec:theory}).
Next, under the common case of $c(x, x)=0$, observe that:

\begin{lemma} $\;$
\label{le:nl_conditions}
\begin{itemize}
\item If $h(x) = \ytilde$, then $h(\decnl_h(x, \ytilde)) = h(x)$
\item If $h(x) \neq \ytilde$, then
$h(\decnl_h(x, \ytilde)) \neq h(x)$ iff $\flipcost_h(x) \le 2$
\end{itemize}
\end{lemma}
Proof in Appendix \ref{pr:nl_conditions}.
Given this, we will define:
\begin{align}
\canflip_h = \{x \,:\, \flipcost_h(x) \le 2\}
\end{align}
to include all points that \emph{can} flip their label.

We can now state our first result.
\begin{theorem} \label{thm:condition1}
Let $D$ be a joint distribution over $\X\times\Y$.
Assume $z=\ytilde=y$ with probability $1-\epsilon$, and $-y$ otherwise,
and assume features are modified via $\decnl$ in Eq. \eqref{eq:response_mapping_nl}.
Then it is better for the system to encourage strategic behavior
iff it learns a classifier $h$ for which it holds that:
\begin{align}
\prob{}{x \not\in \canflip_h \,\wedge\, h(x) \neq y} & +
 \prob{}{x \in \canflip_h} \cdot \epsilon \\
 &\leq \min_{h' \in H} P(h'(x) \neq y) \nonumber
\end{align}
\end{theorem}
The proof (Appendix \ref{pr:condition1})
relies on the coupling of a technical lemma 
with the careful accounting of different error types.
The theorem states that strategic behavior is helpful
if the system can learn a classifier
whose error on points that cannot flip their label
(first summand)
plus an $\epsilon$-proportion of the points that can flip their label (second summand)
is at most the optimal error with no strategic behavior (RHS).
This relates to incentive alignment:
\begin{corollary} \label{cor:nl_cond}
If such an $h$ exists, then the learning task is incentive-aligned.
\end{corollary}
Proof in Appendix \ref{pr:nl_cond}.
Thm. \ref{thm:condition1} states when it is preferable for users to modify features via $\decnl$, rather than simply reporting their original $x$.
For completeness, we also state when 
predictions $\yhat$ on modified inputs are better than $\ytilde$.
\begin{lemma} \label{lem:condition2}
In the same setting as above, it is better for users to use system-provided predictions $\yhat$ than their own side-information $\ytilde$ iff
the system uses a classifier $h$ for which:
\begin{equation}
\frac{\prob{}{x \not\in \canflip_h \,\wedge\, h(x) \neq y}}{\prob{}{x \not\in \canflip_h}} \le \epsilon
\end{equation}
\end{lemma}
Proof in Appendix \ref{pr:condition2}.
Lemma \ref{lem:condition2} states that predictions are helpful as long as
errors on points who cannot flip their label
(and so cannot in principle ``correct'' system errors)
is no more than the noise inherent in side information, $\epsilon$.
Proofs for both results
rely on the coupling of a technical lemma 
with the careful accounting of different error types.
Figure \ref{fig:varying_epsilon} shows on synthetic data
that for most values of $\epsilon$, strategic behavior is preferable
(details in Appendix \ref{sub:varying_exp}).


\subsection{NL as an Extreme Subclass}
Recall that our discussion regarding the need for a
specialize hinge loss for strategic settings
began with the observation that, under a \naive\ approach,
multiple classifiers can obtain a margin of zero.
Our next result shows that, in this sense,
NL is a `extreme' class within problems in GP.
\begin{lemma}
\label{le:zero_margin_set}
Let $S=\{(x_i,y_i)\}_{i=1}^m$. Then of all possible assignments to corresponding side information $\{z_i\}_{i=1}^m$,
the assignment $z_i=y_i$ for all $i \in [m]$
has the largest set of feasible solutions having
zero margin (when such exist).\footnote{Note that the set of feasible solutions may also be large;
the ratio of zero-margin feasible solutions is data-dependent.}
\end{lemma}

\begin{proof}
Let $S=\{(x_i, z_i, y_i)\}_{i=1}^m$ such that $\exists i\in [m], z_i \neq y_i$. 
Denote by $H_{S}'\subseteq H$ the set of all feasible solutions to the naive S-SVM optimization problem (Eq. \ref{eq:naive_ssvm}) for the GP setting.
Denote by $H_{S}''\subseteq H'$ the set of all feasible solutions which induce zero margin $(\min_{i \in [m]}
\left| w^\top \dec_h(x_{i}, z_{i})  \right| = 0)$.

Fix some $i\in[m]$ for which $z_i \neq y_i$.
Consider the sample set $S' = S \setminus \{(x_i, z_i, y_i)\} \cup \{(x_i, y_i, y_i)\}$. 
The $i_{th}$ constraint for $S'$ is strictly weaker in comparison to $S$ (Lemma \ref{lem:s-margin}). The rest of the constraints remain unchanged.
Therefore $\forall h\in H'_S, h\in H_{S'}'$.

If $h \in H_{S}''$ than $\min_{i \in [m]}
\left| w^\top \dec_h(x_{i}, z_{i})  \right| = 0$. This means that $\exists j\in [m], w^\top \dec_h(x_{j}, z_{j}) = 0$. Notice that $j \neq i$. Proof: \\
If $h\in H_{S'}''$ than all constraints must hold. In particular $y_iw^\top x_i < -2$ (Lemma \ref{lem:s-margin}). If $w^\top \dec_h(x_{i}, z_{i}) = 0$ than $h(\dec_h(x_i, z_i)) \neq h(x_i)$ which means that $x_i$ modified it's features to be classified as $z_i$. Since $z_i \neq y_i$, the $i_{th}$ constraint does not hold - a contradiction.

Notice that $w^\top \dec_h(x_{j}, z_{j}) = 0 \Rightarrow \min_{i \in [m]} \left| w^\top \dec_h(x_{i}, z_{i})  \right| = 0 \Rightarrow h\in H_{S'}''$.
We get that $\forall h\in H''_S, h\in H_{S'}''$. Therefore $H''_S \subseteq H_{S'}''$. 
Finally, applying such modifications to $S$ recursively we get that $H''_{S^*}$ is maximal for $S^* = \{(x_i, y_i, y_i)\}_{i=1}^m$.
\end{proof}

\section{Additional Experimental Details}
\paragraph{Technical details.} All experiment results were averaged over $30$ randomized experiments (data generation/shuffles and train/test splits).  In our experiments we used the soft S-SVM for the naive and non-strategic models and used our proposed soft formulation for the strategically aware models. These methods require a hyper-parameter $(\lambda)$ controlling the tradeoff between maximization and amount of constraint violations. We tuned this hyper-parameter using cross-validation of $3$ splits for $\lambda\in \{0.01, 0.1, 1\}$. For the optimization we used the standard Adam optimizer \citep{kingma2017adam} and a learning rate of $0.05$. Each model trained for $200/200/50$ epochs with a batch size of $5/16/24$ for the generalization experiment, the varying preference noise experiment and the PPE experiment respectively.

\subsection{Experiment: Varying $\epsilon$ in NL (Sec. \ref{sec:learning_nl})}
\label{sub:varying_exp}
In this experiment our goal was to emphasize that there are \textbf{simple and common} environments which are incentive-aligned, for both low and high values of $\epsilon$. For this we generated a 2D dataset with clusters sampled from a normal distribution as depicted in Figure \ref{fig:varying_epsilon}. Then generated user preferences according to the Noisy labels setting for various $\epsilon$ values between $0$ and $0.5$. For the learning process we used our proposed approach (Eq. \ref{eq:ss-svm}) and compared its accuracy to the baseline accuracy (non-strategic model) and the pure user-information accuracy $(1-\epsilon)$. Results can be viewed in Figure \ref{fig:varying_epsilon}.

\paragraph{Data.} For this experiment we considered $2$D user features, and the distribution we used is composed out of $4$ sub-distributions:
\begin{itemize}
    \item $(x, y) \sim N\left( \binom{-5}{0}, \binom{0.5 ,\, 0}{0 ,\, 20} \right) \times \{1\text{ w.p. } 0.95, -1 \text{ w.p. } 0.05\}$
    \item $(x, y) \sim N\left( \binom{0.3}{0}, \binom{0.5 ,\, 0}{0 ,\, 20} \right) \times \{1\text{ w.p. } 0, -1 \text{ w.p. } 1\}$
    \item $(x, y) \sim N\left( \binom{-0.3}{0}, \binom{0.5 ,\, 0}{0 ,\, 20} \right) \times \{1\text{ w.p. } 1, -1\text{ w.p. } 0\}$
    \item $(x, y) \sim N\left( \binom{5}{0}, \binom{0.5 ,\, 0}{0 ,\, 20} \right) \times \{1\text{ w.p. } 0.05, -1 \text{ w.p. } 0.95\}$
\end{itemize}
From each distribution we sampled $50$ samples for the train set and $1250$ samples for the test set.

\subsection{Experiment: Generalization (Sec. \ref{sec:experiments_generalization})}
\paragraph{Optimization.} Here we describe how we calculated $\decgp$ (for the naive approach) in a differential way - allowing GD-based optimization methods. Notice that for norm-based cost function, the GP response mapping argmax problem has a closed-form solution:
\begin{equation*}
    \decgp_h(x, z)=\left\{\begin{matrix}
    x- \frac{zw^\top xw}{\norm{w}^2}& -2 \le \frac{zw^\top x}{\norm{w}} \le 0\\
    x & otherwise
\end{matrix}\right.
\end{equation*}
This closed form solution is unfortunately non-differentiable;
instead, we apply the following differentiable approximation:
\begin{equation*}
    \decgp_h(x, z)= x - \frac{zw^\top xw}{\norm{w}^2} \cdot \sigma(cond), \qquad \quad
    cond =(-\frac{zw^\top x}{\norm{w}})\cdot(\frac{zw^\top x}{\norm{w}}+2)
\end{equation*}
Where $\sigma$ is the sigmoid function.

\label{sub:generalization_exp}
\paragraph{Data.} Here we consider 2D user features, and the distributions we used are composed out of two sub-distributions:
\begin{itemize}
    \item[] NL:
    \item $(x, z, y) \sim N\left( \binom{10}{0}, \binom{5 ,\, 0}{0 ,\, 0.2} \right) \times \{ 1 \text{ w.p. } 1\} \times \{1\text{ w.p. } 1\}$
    \item $(x, z, y) \sim N\left( \binom{-10}{0}, \binom{5 ,\, 0}{0 ,\, 0.2} \right) \times \{ -1 \text{ w.p. } 1\} \times \{-1\text{ w.p. } 1\}$
    
    \item[] ADV:
    \item $(x, z, y) \sim N\left( \binom{15.5}{0}, \binom{1.5 ,\, 0}{0 ,\, 0.2} \right) \times \{ -1 \text{ w.p. } 1\} \times \{1\text{ w.p. } 1\}$
    \item $(x, z, y) \sim N\left( \binom{4.5}{0}, \binom{1.5 ,\, 0}{0 ,\, 0.2} \right) \times \{ 1 \text{ w.p. } 1\} \times \{-1\text{ w.p. } 1\}$
    
    \item[] SC:
    \item $(x, z, y) \sim N\left( \binom{15}{0}, \binom{1.5 ,\, 0}{0 ,\, 0.2} \right) \times \{ 1 \text{ w.p. } 1\} \times \{1\text{ w.p. } 1\}$
    \item $(x, z, y) \sim N\left( \binom{4}{0}, \binom{1.5 ,\, 0}{0 ,\, 0.2} \right) \times \{ 1 \text{ w.p. } 1\} \times \{-1\text{ w.p. } 1\}$
    
    \item[] SC (hard):
    \item $(x, z, y) \sim N\left( \binom{2.25}{0}, \binom{0.5 ,\, 0}{0 ,\, 0.2} \right) \times \{ 1 \text{ w.p. } 1\} \times \{1\text{ w.p. } 1\}$
    \item $(x, z, y) \sim N\left( \binom{-2.25}{0}, \binom{0.5 ,\, 0}{0 ,\, 0.2} \right) \times \{ 1 \text{ w.p. } 1\} \times \{-1\text{ w.p. } 1\}$
    
\end{itemize}
From each distribution we sampled $25$ samples for the train set and $1250$ samples for the test set.

\subsection{Experiment: Private Personalized Experiences (PPE) (Sec. \ref{sub:recsys})}
\label{sub:ppe}
Notice that in this setting, $\dec$ does not have a closed-form solution and can only be approximated by \textbf{both} users and the system. This means that even at test time, users will modify their features according to an approximation of $\decppe$. To test the effectiveness and robustness of our proposed algorithm, we fix the approximation of the system and vary the approximation of users.
We trained the system with the MSE loss, and tested $3$ different user approximation losses: MSE loss, classic hinge loss and logistic loss.
For each experiment, we varied the amount of items users were exposed to in advanced (size of $z$). The more items in $z$, the more accurate their "model" will be, and therefore we expect higher accuracy.
\begin{align}
\label{eq:approx_losses}
&\decppetilde_h(x,z) = \argmin_{x' \in \X} \frac{1}{|z|}\sum_{(a, y)\in z} L(x', a, y;W) + \frac{1}{2}\norm{x'-x}^2 &\\
&L^{\mathtt{squared}}(x, a, y;W) = \norm{a^\top Wx-y}^2 & \nonumber \\ 
&L^{\mathtt{hinge}}(x, a, y;W) = \max\{0, 1-y \cdot (a^\top Wx)\} & \nonumber \\ 
&L^{\mathtt{logistic}}(x, a, y;W) = e^{-y \cdot (a^\top Wx)} &\nonumber
\end{align}

\paragraph{System training procedure.}
In PPE the recommender takes $2$ arguments (Eq. \ref{eq:response_mapping_ppe}) so we used an adaptation of the GSC s-hinge loss (Eq.\ref{eq:s-hinge_gsc}) for this experiment.

\begin{align}
\label{eq:s-hinge_ppe}
& \tmplt{\loss}{s-hinge}{PPE}(x,z,a,y;W) \triangleq \\
& \quad\,\, \max\{0, 1 - y\sign(a^\top W \dec_h(x,z))d_\dec(x,z;W)\norm{a^\top W} \} \nonumber
\end{align}

We begin by replacing the $sign$ function with a a sigmoid function - an approximated smooth and differential version of $sign$.
We continue by describing how we calculate $\decppetilde_h(x,z)$. Recall the system uses the $MSE$ approximation and therefore $\decppetilde_h(x,z)$ can be rewritten as:
\begin{align}
\label{eq:response_mapping_ppe2}
&\decppetilde_h(x,z) = \argmin_{x' \in \X} \norm{AWx'-Y}^2 + \frac{|z|}{2}\norm{x'-x}^2 &\\
&A = (a_1, ..., a_{|z|}), Y = (y_1, ..., y_{|z|}) & \nonumber
\end{align}

This is a standard LS with regularization optimization problem and has a closed form solution:
\begin{align}
\label{eq:response_mapping_ppe_sol}
\decppetilde_h(x,z) = (2(AW)^\top AW + |z|\mathbb{I})^{-1}(2(AW)^\top Y + |z|x)
\end{align}

Next, we describe how we find $d_\dec(x,z;W)$.
Notice that according to the MSE approximation, users' strategic modifications are linear transformations:

\begin{align}
\label{eq:response_mapping_ppe_sol_linear}
&\decppetilde_h(x,z) = \alpha_{W, z} x + \beta_{W, z} & \\
&\alpha_{W, z} = |z|(2(AW)^\top AW + |z|\mathbb{I})^{-1} & \nonumber \\
&\beta_{W, z} = (2(AW)^\top AW + |z|\mathbb{I})^{-1}(2(AW)^\top Y) & \nonumber
\end{align}

Being a linear transformation, it is also continuous. This implies that for any $x$, the closest $x'$ to it for which $h(\dec_h(x,z)) \neq h(\dec_h(x',z))$ (Eq. \ref{eq:s-distance}), has to hold $\frac{|aW\dec_h(x',z)|}{\norm{aW}}=0$. Meaning it's distance to the hyperplane after applying modifications is $0$.

Now, we plug in Eq. \ref{eq:response_mapping_ppe_sol_linear} to this constraint and get:
\begin{align}
\label{eq:PPE_constraint}
aW\alpha_{W, z}x - aW\beta_{W, z} = 0
\end{align}

All that remains is to solve a convex optimization problem with linear equality constraints to find $d_\dec(x,z;W)$. To do this in a differential way, we apply another approximation and solve the following optimization problem instead:

\begin{align}
\label{eq:s-distance_PPE}
\decppetilde_h(x,z) = \argmin_{x' \in \X} \norm{aW\alpha_{W, z}x - aW\beta_{W, z}}^2 + \lambda \norm{x'-x}^2
\end{align}

For fairly small values of $\lambda$ (In our experiments we used $\lambda=0.01$). This optimization problem too has a closed form solution which concludes the training procedure calculations.

\section{Additional Illustrations} \label{apdx:illustrations}

\begin{figure}[h!] 
    \centering
    \includegraphics[width=0.3\linewidth]{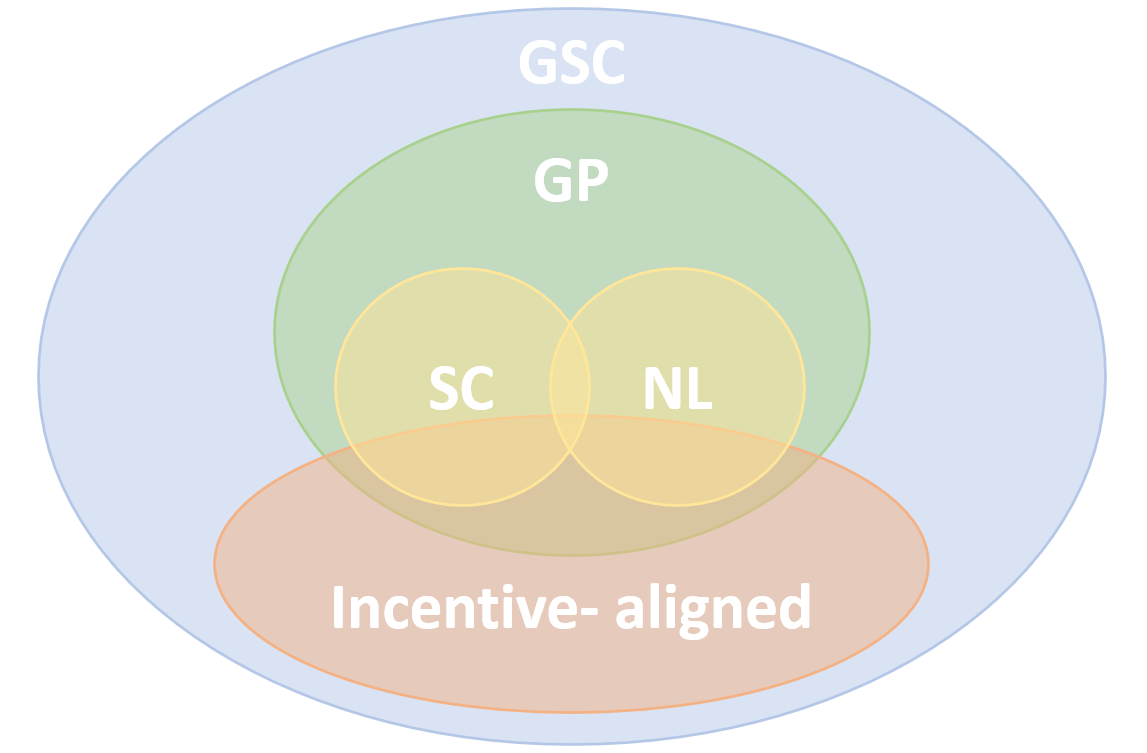} 
    \caption{A depiction of the generalized strategic classification world and various sub-classes within it.}
    \label{fig:SC_wolrd}
\end{figure}

\begin{figure}[h!] 
    \centering
    \includegraphics[width=0.3\linewidth]{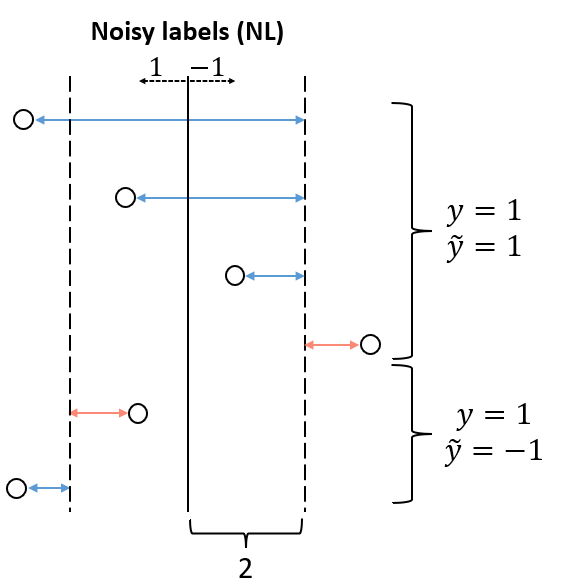} 
    \caption{An illustration of the GP/NL s-margin minimization objective for different samples. The hyperplane classifies points to its left as $-1$ (orange side) and points to its right as $+1$ (blue side). The red and blue lines represent $d_\dec$ - which in this setting is the distances of users to the shifted hyperplane 2 units into their "undesired side". Red lines mean that the user is in a "red" region $(\sign(yw^\top \dec_h(x;z) < 0)$ and similarly for blue lines $(\sign(yw^\top \dec_h(x;z) > 0)$.} 
    \label{fig:s-margin}
\end{figure}

\begin{figure}[h!] 
    \centering
    \includegraphics[width=0.3\linewidth]{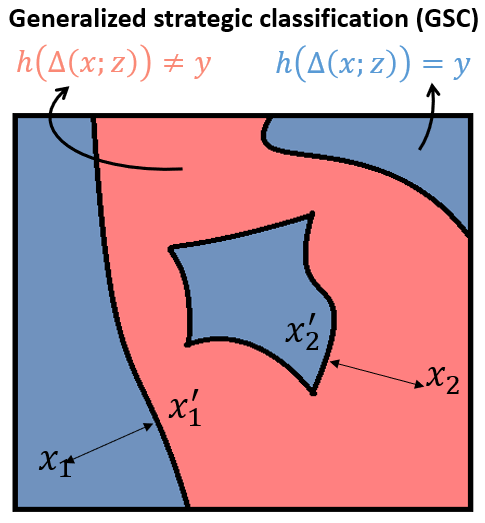} 
    \caption{Let $h, \dec, z, y$ be a classifier, a response mapping, user latent information and a label. Notice that these elements induce regions in $\X$ for which $h(\dec(x;z))=y$ (blue) and $h(\dec(x;z)) \neq y$ (red). The objective of the gs-hinge (Eq. \ref{eq:s-hinge_gsc}) is to maximize $d_\dec(x, z;h)$ (Eq. \ref{eq:s-distance}) for $x$ values in blue regions, and minimize it for $x$ values in red regions. The reason is that $\sign(yw^\top \dec_h(x_1;z))d_{\dec_h}(x_1, z;h) > 0$ and $\sign(yw^\top \dec_h(x_2;z))d_{\dec_h}(x_2, z;h) < 0$.}
    \label{fig:s-loss}
\end{figure}

\section{Additional Definitions}

\subsection{NL Hard formulation equivalent optimization problem}
\label{sub:ssvm}
According to Lemma \ref{lem:s-margin}, the Hard formulation from Eq. \ref{eq:naive_ssvm} for the NL setting can be rewritten as:
\begin{align}
\label{eq:nl_ssvm}
 \argmax_{w\,:\,\norm{w} = 1} \, & \min_{i \in [m]} 
\left| y_i(w^\top x + 2\ytilde_i)  \right| \nonumber \\
& \text{ s.t.}  \,\,\, y_i(w^\top x + 2\ytilde_i) > 0 \quad \forall i \in [m] 
\end{align}
This optimization problem can be reformulated as follows:
\begin{algorithm}[h]
   \caption{NL Hard formulation}
   \label{alg:ssvm}
\begin{algorithmic}
   \STATE {\bfseries Input:} $\{(x_i, \ytilde_i, y)\}^m$
   \STATE {\bfseries Solve:}
        \STATE $w_0 = \argmin_{w} \norm{w}^2$
        \STATE s.t. $\forall i\in [m], y_i(w^\top x + 2\ytilde_i\norm{w}) \ge 1$
   \STATE {\bfseries Output:} $\hat{w} = \frac{w_0}{\norm{w_0}}$
\end{algorithmic}
\end{algorithm}

Algorithm \ref{alg:ssvm} and Eq. \ref{eq:nl_ssvm} output the same classifier $w$. Proof in Appendix \ref{pr:ssvm}.

\subsection{NL/GP Soft formulation} \label{sub:soft_ssvm}
The hard constraints of the algorithm \ref{alg:ssvm} optimization problem can be soften by introducing slack variables, similarly to the classic soft SVM algorithm.
\begin{algorithm}[h]
   \caption{NL/GP Soft formulation}
   \label{alg:soft_ssvm}
\begin{algorithmic}
   \STATE {\bfseries Input:} $\{(x_i, \ytilde_i, y)\}^m$
   \STATE {\bfseries Parameter:} $\lambda > 0$
   \STATE {\bfseries Solve:}
        \STATE $w, \xi_i = \argmin_{w, \xi_i} \lambda \norm{w}^2 + 
        \frac{1}{m}\sum_{i=1}^m \xi_i$
        \STATE s.t. $\forall i\in [m], y_i(w^\top x + 2\ytilde_i\norm{w}) \ge 1-\xi_i$ and $\xi_i \ge 0$
   \STATE {\bfseries Output:} $w$
\end{algorithmic}
\end{algorithm}

We can rewrite the optimization problem of Algorithm \ref{alg:soft_ssvm} as a regularized loss-minimization problem:

\begin{align*}
    &w, \xi_i = \argmin_{w, \xi_i} \lambda \norm{w}^2 + \frac{1}{m}\tmplt{\loss}{s-hinge}{NL}(x_i,\ytilde_i,y_i;w) & \\
    &\tmplt{\loss}{s-hinge}{NL}(x,\ytilde,y;w) \triangleq \max\{ 0, 1-y(w^\top x + 2 \ytilde \norm{w}) \} &
\end{align*}

Proof in Appendix \ref{pr:soft_ssvm}.

\section{Proofs} \label{sec:proofs}

\begingroup
\allowdisplaybreaks

\subsection{Lemma \ref{le:nl_conditions}}
\label{pr:nl_conditions}
\begin{proof}
Assume $c(x, x)=0$.
\begin{itemize}
\item If $h(x) = \ytilde$: Assume, for sake of contradiction, that $h(\decnl_h(x, \ytilde)) \neq h(x)$. This implies $h(\decnl_h(x, \ytilde)) = -\ytilde$. 
\item 
Recall the definition:
\[
\decnl_h(x, \ytilde) = \argmax_{x' \in \X} \1{h(x')=\ytilde} - \frac{1}{2}c(x,x')
\]
This means that $\1{h(x)=\ytilde} - \frac{1}{2}c(x,x) \le \1{h(\decnl_h(x, \ytilde))=\ytilde} - \frac{1}{2}c(x,\decnl_h(x, \ytilde))$.
But $\1{h(x)=\ytilde} - \frac{1}{2}c(x,x) = 1$; hence, since:
\[
\1{h(\decnl_h(x, \ytilde))=\ytilde} - \frac{1}{2}c(x,\decnl_h(x, \ytilde)) = \1{-\ytilde=\ytilde} - \frac{1}{2}c(x,\decnl_h(x, \ytilde)) = - \frac{1}{2}c(x,\decnl_h(x, \ytilde)) \le 0 < 1
\]
we get that:
\[
\1{h(x)=\ytilde} - \frac{1}{2}c(x,x) > \1{h(\decnl_h(x, \ytilde))=\ytilde} - \frac{1}{2}c(x,\decnl_h(x, \ytilde))
\]
Therefore the assumption must be false.

\item If $h(x) \neq \ytilde$: 
$h(\decnl_h(x, \ytilde)) \neq h(x)$ iff $\flipcost_h(x) \le 2$
\begin{itemize}
    \item If $h(\decnl_h(x, \ytilde)) \neq h(x)$: \\
    So $h(\decnl_h(x, \ytilde)) = \ytilde$. \\
    Assume, for sake of contradiction, that $\flipcost_h(x) > 2$. \\
    Therefore $c(x, \decnl_h(x, \ytilde)) > 2$. \\
    Again, $\1{h(x)=\ytilde} - \frac{1}{2}c(x,x) \le \1{h(\decnl_h(x, \ytilde))=\ytilde} - \frac{1}{2}c(x,\decnl_h(x, \ytilde))$. \\
    But \\
    $\1{h(x)=\ytilde} - \frac{1}{2}c(x,x) = 0$. \\
    $\1{h(\decnl_h(x, \ytilde))=\ytilde} - \frac{1}{2}c(x,\decnl_h(x, \ytilde)) = \1{\ytilde=\ytilde} - \frac{1}{2}c(x,\decnl_h(x, \ytilde)) < 1 - \frac{1}{2}\cdot2 = 0$. \\
    So $\1{h(x)=\ytilde} - \frac{1}{2}c(x,x) > \1{h(\decnl_h(x, \ytilde))=\ytilde} - \frac{1}{2}c(x,\decnl_h(x, \ytilde))$. \\
    Therefore the assumption must be false.
    
    \item If $\flipcost_h(x) \le 2$: \\
    Let $x'$ be the minimizer of the flipping cost optimization problem (Definition \ref{def:flipping_cost}). \\
    Therefore $h(x) \neq h(x')$ and $c(x, x') \le 2$. This means that $h(x')=\ytilde$. \\
    Assume, for sake of contradiction, that $h(\decnl_h(x, \ytilde)) = h(x)$. \\
    So $h(\decnl_h(x, \ytilde)) = -\ytilde$. \\
    Again, $\1{h(x')=\ytilde} - \frac{1}{2}c(x,x') \le \1{h(\decnl_h(x, \ytilde))=\ytilde} - \frac{1}{2}c(x,\decnl_h(x, \ytilde))$. \\
    But \\
    $\1{h(x')=\ytilde} - \frac{1}{2}c(x,x') = \1{\ytilde=\ytilde} - \frac{1}{2}c(x,x') \ge 1-\frac{1}{2}\cdot2 = 0 > -1$. \\
    $\1{h(\decnl_h(x, \ytilde))=\ytilde} - \frac{1}{2}c(x,\decnl_h(x, \ytilde)) = \1{-\ytilde=\ytilde} - \frac{1}{2}c(x,\decnl_h(x, \ytilde)) = -1 - \frac{1}{2}c(x,\decnl_h(x, \ytilde)) \le 0$. \\
    So $\1{h(x)=\ytilde} - \frac{1}{2}c(x,x) > \1{h(\decnl_h(x, \ytilde))=\ytilde} - \frac{1}{2}c(x,\decnl_h(x, \ytilde))$. \\
    Therefore the assumption must be false.
\end{itemize}
\end{itemize}
\end{proof}

\subsection{Theorem \ref{thm:condition1}}
\label{pr:condition1}
\begin{proof}
Let $D$ be a joint distribution over $\X\times\Y$.
Assume $z=\ytilde=y$ with probability $1-\epsilon$, and $-y$ otherwise,
and assume features are modified via $\decnl$ in Eq. \eqref{eq:response_mapping_nl}.
It is better for the system to encourage strategic behavior
iff it learns a classifier $h$ for which it holds that:
\begin{equation*}
\mathbb{E}_{D}[\1{h(\decnl_h(x;\ytilde)) \neq y}] \leq
\min_{h' \in H} \mathbb{E}_{D}[\1{h'(x) \neq y}]
\end{equation*}

\begin{itemize}
    \item RHS: We have
    $\mathbb{E}_{D}[\1{h'(x) \neq y}] = \prob{}{h'(x) \neq y} $
    
    \item LHS: 
    From Lemma \ref{le:nl_conditions} we derive two conclusions:
    \begin{itemize}
        \item 1. $x\in \canflip_h \Rightarrow h(\decnl_h(x;\ytilde))=\ytilde$.
        \item 2. $x\notin \canflip_h \Rightarrow h(\decnl_h(x;\ytilde))=h(x)$.
    \end{itemize}
    This gives us:
    \begin{flalign*}
    &\mathbb{E}_{D}[\1{h(\decnl_h(x;\ytilde)) \neq y}] = \prob{}{h(\decnl_h(x;\ytilde)) \neq y} = &\\
    &\prob{}{h(\decnl_h(x;\ytilde)) \neq y | x\in \canflip_h}\cdot \prob{}{x\in \canflip_h} + \prob{}{h(\decnl_h(x;\ytilde)) \neq y | x\notin \canflip_h}\cdot \prob{}{x\notin \canflip_h} =_{1, 2} &\nonumber \\
    &\prob{}{\ytilde \neq y | x\in \canflip_h}\cdot \prob{}{x\in \canflip_h} + \prob{}{h(x) \neq y | x\notin \canflip_h}\cdot \prob{}{x\notin \canflip_h} =_{\ytilde \text{ and } x \text{ are i.i.d}} & \nonumber \\
   &\epsilon\cdot \prob{}{x\in \canflip_h} + \prob{}{h(x) \neq y \wedge x\notin \canflip_h} &\nonumber
    \end{flalign*}
\end{itemize}

Finally, we get:
\begin{align*}
&\mathbb{E}_{D}[\1{h(\decnl_h(x;\ytilde)) \neq y}] \leq
\min_{h' \in H} \mathbb{E}_{D}[\1{h'(x) \neq y}] \\
& \Leftrightarrow 
\epsilon\cdot \prob{}{x\in \canflip_h} + \prob{}{h(x) \neq y \wedge x\notin \canflip_h} \leq
\min_{h' \in H} \prob{}{h'(x) \neq y}&
\end{align*}
\end{proof}

\subsection{Corollary \ref{cor:nl_cond}}
\label{pr:nl_cond}
\begin{proof}
if such an $h$ exists, then according to Theorem \ref{thm:condition1}:
\begin{equation*}
\mathbb{E}_{D}[\1{h(\decnl_h(x;\ytilde)) \neq y}] \leq
\min_{h' \in H} \mathbb{E}_{D}[\1{h'(x) \neq y}]
\end{equation*}
Therefore $\exists h \in H$ such that $\forall h'\in H$ it holds that:
\begin{equation*}
\mathbb{E}_{D}[\1{h(\decnl_h(x;\ytilde)) \neq y}] \leq
\mathbb{E}_{D}[\1{h'(x) \neq y}]
\end{equation*}
According to Definition \ref{def:ia}, the environment is incentive-aligned.
\end{proof}

\subsection{Lemma \ref{lem:condition2}}
\label{pr:condition2}
\begin{proof}
Let $D$ be a joint distribution over $\X\times\Y$.
Assume $z=\ytilde=y$ with probability $1-\epsilon$, and $-y$ otherwise,
and assume features are modified via $\decnl$ in Eq. \eqref{eq:response_mapping_nl}.
It is better for users to use system-provided predictions $\yhat$ than their own side-information $\ytilde$ iff the system learns a classifier $h$ for which it holds that:
\begin{equation*}
\mathbb{E}_{D}[\1{h(\decnl_h(x;\ytilde)) \neq y}] \leq
\mathbb{E}_{D}[\1{\ytilde \neq y}]
\end{equation*}

\begin{itemize}
    \item RHS: We have $\mathbb{E}_{D}[\1{\ytilde \neq y}] = \prob{}{\ytilde \neq y} = \epsilon$.
    
    \item LHS: \\
    \begin{align*}
    &\mathbb{E}_{D}[\1{h(\decnl_h(x;\ytilde)) \neq y}] = \prob{}{h(\decnl_h(x;\ytilde)) \neq y} =_{\text{Theorem \ref{thm:condition1} proof}} \\
   &\epsilon\cdot \prob{}{x\in \canflip_h} + \prob{}{h(x) \neq y \wedge x\notin \canflip_h} 
    \end{align*}
\end{itemize}

Finally, we get:
\begin{align*}
&\mathbb{E}_{D}[\1{h(\decnl_h(x;\ytilde)) \neq y}] \leq
\min_{h' \in H} \mathbb{E}_{D}[\1{h'(x) \neq y}] \\
& \Leftrightarrow 
\epsilon\cdot \prob{}{x\in \canflip_h} + \prob{}{\ytilde \neq y \wedge x\notin \canflip_h} \leq
\epsilon \\
& \Leftrightarrow 
\prob{}{\ytilde \neq y \wedge x\notin \canflip_h} \leq
\epsilon(1 - \prob{}{x\in \canflip_h}) \\
& \Leftrightarrow 
\prob{}{\ytilde \neq y \wedge x\notin \canflip_h} \leq
\epsilon \cdot \prob{}{x\notin \canflip_h} \\
& \Leftrightarrow
\frac{\prob{}{\ytilde \neq y \wedge x\notin \canflip_h}}{\prob{}{x\notin \canflip_h}} \le \epsilon 
\end{align*}
\end{proof}

\subsection{Lemma \ref{lem:s-margin}}
\label{pr:s-margin}
\begin{proof}
Let $h = (w)$ be a linear classifier and $(x, \ytilde, y)\in \X \times Y \times Y$. We show that:
\begin{equation*}
y w^\top \decnl_h(x,\ytilde) >0
\,\,\Leftrightarrow \,\,
y(w^\top x + 2\ytilde\norm{w}) >0
\end{equation*}

Or equivalently:

\begin{equation*}
h(\decnl_h(x,\ytilde)) = y
\,\,\Leftrightarrow \,\,
\frac{yw^\top x}{\norm{w}} > -2\ytilde y
\end{equation*}

Observation:
In $NL$, the cost function is the euclidean distance between two vectors $(c(x, x')=\norm{x-x'})$. Therefore the flipping cost (Definition \ref{def:flipping_cost}) of a user $x$ is the euclidean distance to the separating hyperplane $(\flipcost_h(x) = \frac{|w^\top x|}{\norm{w}})$.

\begin{itemize}
    \item If $y=\ytilde$:
    \begin{itemize}
        \item If $h(x) = \ytilde$: From Lemma \ref{le:nl_conditions},
        we have $h(\decnl_h(x,\ytilde)) = \ytilde = y$. Hence,
        \[
        yw^\top x =\ytilde w^\top x > 0 \Leftrightarrow \frac{yw^\top x}{\norm{w}} > 0 > -2 = -2\ytilde y
        \]
        
        \item If $h(x) \neq \ytilde$: From Lemma \ref{le:nl_conditions},
        we have $h(\decnl_h(x,\ytilde)) = \ytilde = y \Leftrightarrow \flipcost_h(x) \le 2$. Hence,
        \begin{align*}
        & \flipcost_h(x) \le 2 \Leftrightarrow \frac{|w^\top x|}{\norm{w}} \le 2 \qquad \text{(observation)} \\
        & \frac{|w^\top x|}{\norm{w}} \le 2 \Leftrightarrow \frac{\ytilde w^\top x}{\norm{w}} \ge -2 \qquad (\ytilde w^\top x < 0) \\
        & \frac{\ytilde w^\top x}{\norm{w}} \ge -2 \Leftrightarrow \frac{yw^\top x}{\norm{w}} \ge -2\ytilde y
        \end{align*}
    \end{itemize}   
    
    \item If $y \neq \ytilde$:
    \begin{itemize}
        \item If $h(x) = \ytilde$: From Lemma \ref{le:nl_conditions},
        we have $h(\decnl_h(x,\ytilde)) = \ytilde = -y$. Hence,
        \[
        yw^\top x = -\ytilde w^\top x < 0 \Leftrightarrow \frac{yw^\top x}{\norm{w}} < 0 < 2 = -2\ytilde y
        \]
        
        \item If $h(x) \neq \ytilde$:
        From Lemma \ref{le:nl_conditions},
        we have $h(\decnl_h(x,\ytilde)) = \ytilde = -y \Leftrightarrow \flipcost_h(x) \le 2$. Hence,
        \begin{align*}
        & \flipcost_h(x) \le 2 \Leftrightarrow \frac{|w^\top x|}{\norm{w}} \le 2 \qquad \text{(observation)} \\
        & \frac{|w^\top x|}{\norm{w}} \le 2 \Leftrightarrow \frac{\ytilde w^\top x}{\norm{w}} \ge -2 \qquad (\ytilde w^\top x < 0) \\
        & \frac{\ytilde w^\top x}{\norm{w}} \ge -2 \Leftrightarrow \frac{yw^\top x}{\norm{w}} < -2\ytilde y
        \end{align*}
    \end{itemize}
\end{itemize}
\end{proof}

\subsection{Hard formulation equivalent form (Appendix \ref{sub:ssvm})}
\label{pr:ssvm}
\begin{proof}
We begin by rewriting Eq. \ref{eq:nl_ssvm}:
\begin{equation}
\label{eq:ssvm_alt}
    \argmax_{w\,:\,\norm{w} = 1} \, \min_{i \in [m]} 
    y_i(w^\top x_{i} +2 \ytilde_i)
\end{equation}

The two are equivalent whenever there is a solution to the preceding problem, i.e. the data is strategically separable (Definition \ref{def:strategic_separability}).

Let $w^*$ be a solution of Eq. \ref{eq:ssvm_alt}. Define $\gamma^* = \min_{i \in [m]} 
    y_i(w^\top x_{i} +2 \ytilde_i)$. Therefore $\forall i\in [m]$ we have:
\begin{equation*}
    y_i((w^*)^\top x_{i} +2 \ytilde_i) \ge \gamma^*
\end{equation*}

Or equivalently (since $\gamma^* > 0$ and $\norm{w^*} = 1$):

\begin{equation*}
    y_i((\frac{w^*}{\gamma^*})^\top x_{i} +2 \ytilde_i \norm{\frac{w^*}{\gamma^*}}) \ge 1
\end{equation*}

Hence, $\frac{w^*}{\gamma^*}$ satisfies the conditions of the optimization in Algorithm \ref{alg:ssvm}. Therefore, $\norm{w_0} \ge \norm{\frac{w^*}{\gamma^*}} = \frac{1}{\gamma^*}$.

It follows that $\forall i\in [m]$:
\begin{equation*}
    y_i(\hat{w}^\top x_{i} +2 \ytilde_i) = 
    \frac{1}{\norm{w_0}}y_i(w_0^\top x_{i} +2 \ytilde_i\norm{w_0}) \ge \frac{1}{\norm{w_0}} \ge \gamma^*
\end{equation*}

Since $\norm{\hat{w}} = 1$ we obtain that $\hat{w}$ is an optimal solution of Eq. \ref{eq:ssvm_alt}.
\end{proof}

\subsection{Soft formulation equivalent form (Appendix \ref{sub:soft_ssvm})}
\label{pr:soft_ssvm}
\begin{proof}
Fix some $w, i$ and consider the minimization over $\xi_i$ in Algorithm \ref{alg:soft_ssvm}.
Since $\xi_i$ must be nonnegative, the best assignment to $x_i$ would be $0$
if $y_i(w^\top x_i + 2\ytilde_i\norm{w}) \ge 1$ and would be $1 - y_i(w^\top x_i + 2\ytilde_i\norm{w})$ otherwise. Therefore
\begin{equation*}
    \xi_i = \left\{\begin{matrix}
0, & y_i(w^\top x_i + 2\ytilde_i\norm{w}) \ge 1\\ 
1-y_i(w^\top x_i + 2\ytilde_i\norm{w}), & y_i(w^\top x_i + 2\ytilde_i\norm{w}) < 1 
\end{matrix}\right. = \max\{ 0, 1-y_i(w^\top x_i + 2 \ytilde_i \norm{w}) \}
\end{equation*}
\end{proof}

\subsection{Definition \ref{def:s-hinge} equivalent form}
\label{pr:s-hinge}
\begin{proof}
We will show that for the $NL$ setting:
\begin{equation*}
\tmplt{\loss}{s-hinge}{NL}(x,\ytilde,y;w) \triangleq
\max\{ 0, 1-y(w^\top x + 2 \ytilde \norm{w}) \} = \max \{ 0, 1 - yw^\top \decnl_h(x,\ytilde) - 
(2-c(x,\decnl_h(x,\ytilde)))y\ytilde\norm{w}\}
\end{equation*}

Assumption: \\
Users do not change their features unless the change strictly increase their utility, i.e. $h(\decnl_h(x;\ytilde)) = h(x) \Rightarrow \decnl_h(x;\ytilde) = x$.

Observation: \\
Assuming $c(x, x') = \norm{x-x'}$ as in NL,
the closest point $x'$ to $x$ for which $h(x') \neq h(x)$ is a point on the hyperplane $w$ closest to $x$. The modification cost is the euclidean distance between $x$ and the hyperplane $(c(x, x') = \frac{|w^\top x|}{\norm{w}})$

Let $h = (w)$ be a linear classifier and $(x, \ytilde, y)\in \X \times Y \times Y$. 
\begin{itemize}
    \item If $\frac{\ytilde w^\top x}{\norm{w}} > 0$: \\
    $\ytilde w^\top x > 0 \Rightarrow$ \\
    $h(x) = \ytilde \Rightarrow$  \qquad (Lemma \ref{le:nl_conditions}). \\
    $h(\decnl_h(x;\ytilde)) = h(x) = \ytilde \Rightarrow$ \qquad (assumption).\\
    $\decnl_h(x;\ytilde) = x \Rightarrow$ \\
    $yw^\top \decnl_h(x;\ytilde) = yw^\top x \Rightarrow$ \qquad $(c(x, \decnl_h(x;\ytilde)) = c(x,x) = 0)$ \\
    $\max\{ 0, 1-y(w^\top x + 2 \ytilde \norm{w}) \} = \max \{ 0, 1 - yw^\top \decnl_h(x,\ytilde) - 
    (2-c(x,\decnl_h(x,\ytilde)))y\ytilde\norm{w}\}$
    
    \item If $0 \ge \frac{\ytilde w^\top x}{\norm{w}} \ge -2$: \\
    $\ytilde w^\top x < 0 \Rightarrow$ \\
    $h(x) \neq \ytilde \Rightarrow$  \qquad (Lemma \ref{le:nl_conditions}). \\
    $(h(\decnl_h(x;\ytilde)) \neq h(x) \Leftrightarrow \flipcost_h(x) \le 2) \Rightarrow$ \qquad $(\flipcost_h(x) = \frac{|w^\top x|}{\norm{w}} \text{( see observation). Therefore } \flipcost_h(x) \le 2)$.\\
    $h(\decnl_h(x;\ytilde)) \neq h(x) \Rightarrow$ \qquad (observation). \\
    $yw^\top \decnl_h(x;\ytilde) = 0 \wedge c(x, \decnl_h(x;\ytilde)) = \frac{|w^\top x|}{\norm{w}} \Rightarrow$ \qquad $(\frac{|w^\top x|}{\norm{w}} = \frac{-\ytilde w^\top x}{\norm{w}})$. \\
    $\max\{ 0, 1-y(w^\top x + 2 \ytilde \norm{w}) \} = \max \{ 0, 1 - yw^\top \decnl_h(x,\ytilde) - 
    (2-c(x,\decnl_h(x,\ytilde)))y\ytilde\norm{w}\}$
    
    \item If $\frac{\ytilde w^\top x}{\norm{w}} < -2$:
    $\ytilde w^\top x < 0 \Rightarrow$ \\
    $h(x) \neq \ytilde \Rightarrow$  \qquad (Lemma \ref{le:nl_conditions}). \\
    $(h(\decnl_h(x;\ytilde)) \neq h(x) \Leftrightarrow \flipcost_h(x) \le 2) \Rightarrow$ \qquad $(\flipcost_h(x) = \frac{|w^\top x|}{\norm{w}} \text{ (see Lemma \ref{lem:s-margin} proof). Therefore } \flipcost_h(x) > 2)$.\\
    $h(\decnl_h(x;\ytilde)) = h(x) \Rightarrow$ \qquad (assumption). \\
    $\decnl_h(x;\ytilde) = x \Rightarrow$ \\
    $yw^\top \decnl_h(x;\ytilde) = yw^\top x \Rightarrow$ \qquad $(c(x, \decnl_h(x;\ytilde)) = c(x,x) = 0)$ \\
    $\max\{ 0, 1-y(w^\top x + 2 \ytilde \norm{w}) \} = \max \{ 0, 1 - yw^\top \decnl_h(x,\ytilde) - 
    (2-c(x,\decnl_h(x,\ytilde)))y\ytilde\norm{w}\}$
\end{itemize}
\end{proof}

\subsection{Strategic distance for GP}
\label{pr:flipping_distance_gp}
Claim: $\tmplt{d}{\mathnormal{\dec}}{GP}(x,z;w)=
| \wbar^\top x + 2z |$
\begin{proof}
Let $h = (w)$ be a linear classifier, $x, x'\in \X$ and $z\in \Z$. We show necessary and sufficient conditions for which $h(\decgp_h(x)) \neq h(\decgp_h(x'))$.

\begin{itemize}
    \item If $h(x) = z$: \\
    $h(\decgp_h(x)) \neq h(\decgp_h(x')) \Leftrightarrow$ \qquad (Lemma \ref{le:nl_conditions}). \\
    $z \neq h(\decgp_h(x')) \Leftrightarrow$ \qquad (Lemma \ref{le:nl_conditions}). \\
    $h(x') \neq z \wedge \flipcost_h(x') > 2 \Leftrightarrow$ \\
    $zw^\top x' < 0 \wedge \frac{|w^\top x'|}{\norm{w}} > 2 \Leftrightarrow$\\
    $\frac{zw^\top x'}{\norm{w}} < -2$. \\
    Also, $zw^\top x > 0$. Therefore $x$ and any $x'$ for which $\frac{zw^\top x'}{\norm{w}} < -2$ are on different sides of the hyperplane.\\
    The minimal distance from $x$ to such $x'$ is the distance from $x$ to the hyperplane plus $2$:\\
    $\tmplt{d}{\mathnormal{\dec}}{GP}(x,z;w) = \frac{|w^\top x|}{\norm{w}} + 2 = \frac{zw^\top x}{\norm{w}} + 2 = |\frac{w^\top x}{\norm{w}} + 2z| = | \wbar^\top x + 2z |$
    
    \item If $h(x) \neq z$: \\
    \begin{itemize}
        \item If $\flipcost_h(x) \le 2$: \\
        $h(\decgp_h(x)) \neq h(\decgp_h(x')) \Leftrightarrow$ \qquad (Lemma \ref{le:nl_conditions}). \\
        $z \neq h(\decgp_h(x')) \Leftrightarrow$ \qquad (Lemma \ref{le:nl_conditions}). \\
        $h(x') \neq z \wedge \flipcost_h(x') > 2 \Leftrightarrow$ \\
        $zw^\top x' < 0 \wedge \frac{|w^\top x'|}{\norm{w}} > 2 \Leftrightarrow$\\
        $\frac{zw^\top x'}{\norm{w}} < -2$. \\
        Also, $zw^\top x \le 0$ and $\frac{|w^\top x'|}{\norm{w}} \le 2$.\\
        This means that $-2 \le \frac{zw^\top x'}{\norm{w}} \le 0$. \\
        Therefore $x$ and any $x'$ for which $\frac{zw^\top x'}{\norm{w}} < -2$ are on the same side of the hyperplane and $x$ is closer to the hyperplane than $x'$.\\
        The minimal distance from $x$ to such $x'$ is $2$ minus the distance from $x$ to the hyperplane:\\
        $\tmplt{d}{\mathnormal{\dec}}{GP}(x,z;w) = 2 - \frac{|w^\top x|}{\norm{w}} = 2 + \frac{zw^\top x}{\norm{w}} = |\frac{w^\top x}{\norm{w}} + 2z| = | \wbar^\top x + 2z |$
        
        \item If $\flipcost_h(x) > 2$: \\
        $h(\decgp_h(x)) \neq h(\decgp_h(x')) \Leftrightarrow$ \qquad (Lemma \ref{le:nl_conditions}). \\
        $z = h(\decgp_h(x')) \Leftrightarrow$ \qquad (Lemma \ref{le:nl_conditions}). \\
        $h(x') = z \vee (h(x') \neq z \wedge \flipcost_h(x') \le 2) \Leftrightarrow$ \\
        $zw^\top x' > 0 \vee (zw^\top x' \le 0 \wedge \frac{|w^\top x'|}{\norm{w}} \le 2) \Leftrightarrow$\\
        $\frac{zw^\top x'}{\norm{w}} > 0 \vee (-2 \le \frac{zw^\top x'}{\norm{w}} \le 0) \Rightarrow$ \\
        $\frac{zw^\top x'}{\norm{w}} \ge -2.$ \\
        Also, $zw^\top x \le 0$ and $\frac{|w^\top x'|}{\norm{w}} > 2$.\\
        This means that $\frac{zw^\top x'}{\norm{w}} < -2$. \\
        The minimal distance from $x$ to an $x'$ for which $\frac{zw^\top x'}{\norm{w}} \ge -2$ is the distance from $x$ to the hyperplane minus $2$:\\
        $\tmplt{d}{\mathnormal{\dec}}{GP}(x,z;w) = \frac{|w^\top x|}{\norm{w}} - 2 = -\frac{zw^\top x}{\norm{w}} - 2 = |\frac{w^\top x}{\norm{w}} + 2z| = | \wbar^\top x + 2z |$.
    \end{itemize}
\end{itemize}
\end{proof}

\subsection{Theorem \ref{thm:bound_gsc}}
\label{pr:bound_gsc}
\begin{proof}
Denote by $s$ a sample $(x, z, y) \in \X \times \Z \times \Y$.
Recall:
\begin{align*}
    &\tmplt{\Loss}{0/1}{GSC}(s;w) \triangleq \1{y\sign(w^\top \dec_h(x;z))}& \\
    &\tmplt{\loss}{s-hinge}{GSC}(s;w) \triangleq \max\{0, 1 - y\sign(w^\top \dec_h(x,z))d_\dec(x,z;w)\norm{w} \}&
\end{align*}
Since $d_\dec(x,z;w) \ge 0$ it is clear that $\tmplt{\Loss}{0/1}{GSC}(s;w) \le \tmplt{\loss}{s-hinge}{GSC}(s;w)$.\\

We rewrite $\tmplt{\loss}{s-hinge}{GSC}(s;w)$:
\begin{align*}
    &\tmplt{\loss}{s-hinge}{GSC}(s;w) \triangleq \max\{0, 1 - y\sign(w^\top \dec_h(x,z))\norm{x-x^d}\norm{w}\}& \\
    &\text{where}& \\
     & x^d \triangleq \argmin_{x'} \norm{x-x'}& \\
    & \qquad \text{ s.t. } \,\, h(\dec_h(x,z)) \neq h(\dec_h(x',z))&
\end{align*}

Define $\H_k = \{s \rightarrow y\sign(w^\top \dec_h(x,z))\norm{x-x_d}\norm{w} : \norm{w} \le k \}$ and let $S=\{s_i\}^m$ be vectors in that space. Denote $r=\max_{x \in \X}$. \\
We bound the Rademacher complexity of $\H_k\circ S$:
\begin{align*}
    &mR(\H_k\circ S) = \mathbb{E}_{\sigma}\left[\sup_{w\in \H_k}\sum_{i=1}^m \sigma_i y_i\sign(w^\top \dec_h(x_i,z_i))\norm{x_i-x^d_i}\norm{w} \right] =_1 & \\
    &\mathbb{E}_{\sigma}\left[\sup_{w\in \H_k}\sum_{i=1}^m \sigma_i \norm{x_i-x^d_i}\norm{w}\right] = \mathbb{E}_{\sigma}\left[\sup_{w\in \H_k} \norm{w}\sum_{i=1}^m \sigma_i \norm{x_i-x^d_i}\right] \le & \\
    &\mathbb{E}_{\sigma}\left[k \left |\sum_{i=1}^m \sigma_i \norm{x_i-x^d_i} \right | \right] =k\mathbb{E}_{\sigma}\left[\sqrt{\left (\sum_{i=1}^m \sigma_i \norm{x_i-x^d_i}\right)^2} \right] \le_2 & \\
    &k\sqrt{\mathbb{E}_{\sigma}\left[\left (\sum_{i=1}^m \sigma_i \norm{x_i-x^d_i}\right)^2 \right]} =k\sqrt{\mathbb{E}_{\sigma}\left[\sum_{i,j} \sigma_i\sigma_j \norm{x_i-x^d_i}\norm{x_j-x^d_j} \right]} =_3& \\
    &k\sqrt{\sum_{i=1}^m\mathbb{E}_{\sigma}\left[ \sigma_i^2 \right] \norm{x_i-x^d_i}^2} \le k\sqrt{m}\cdot \max_i\norm{x_i-x^d_i} \le k\sqrt{m}\cdot \max_i\norm{x_i} + \norm{x^d_i} \le k\sqrt{m}\cdot 2r&
\end{align*}
With steps following from:
\begin{enumerate}
\item $\sigma_i y_i\sign(w^\top \dec_h(x_i,z_i)) \equiv_d \sigma_i$ under the expectancy
\item Jensen's inequality
\item Independence
\end{enumerate}
Therefore. $R(\H_k\circ S) \le \frac{2kr}{\sqrt{m}}$.

Consider the hinge function $\phi(t) = \max\{0, 1-t\}$ for scalar values $t\in \mathbb{R}$. $\phi$ is 1-Lipschitz:
\begin{equation*}
    \forall t_1, t_2 \in \mathbb{R}, \qquad
    \left | \phi(t_1) - \phi(t_2) \right | = \left | \max{0, 1-t_1} - \max{0, 1-t_2} \right | \le \max{|0-0|, |1-t_1 - 1+t_2|} = |t_1-t_2|
\end{equation*}

Then, according to Talagrand’s contraction principal, $R(\phi \circ \H_k) \le R(\H_k)$. \\

Let $D$ be a distribution on $\X \times \Z \times \Y$ such that there exists some $w^*$ with $\prob{D}{y\sign((w^*)^\top \dec_h(x,z) \ge 1} = 1$. Let $w_S$ be the output of the Hard formulation algorithm without normalization:
\begin{align*}
    &w = \argmin_{w} \norm{w}^2& \\
    &s.t. \,\, \forall i\in [m], y_i\sign(w^\top \dec_h(x_i,z_i))\norm{x_i-x^d_i}\norm{w} \ge 1&
\end{align*}
Define $\H^* = \{w: \norm{w} \le \norm{w^*}\}$. We have $\norm{w_S} \le \norm{w^*}$ and therefore $w_S \in \H^*$. From the generalization theorem on Rademacher complexity, with probability greater or equal to $1-\delta$, for all $w \in \H^*$:
\begin{equation*}
    \tmplt{\Loss}{s-hinge}{GSC}(w;D) - \tmplt{\Loss}{s-hinge}{GSC}(w;S) \le 2R_D(\phi \circ \H_{\norm{w^*}}) + c\sqrt{\frac{2ln(\frac{2}{\delta})}{m}}
\end{equation*}
where $c$ is the maximal loss, which in our case is $(1+2r\norm{w^*})$.

Finally, we get:
\begin{equation*}
    \tmplt{\Loss}{s-hinge}{GSC}(w;D) \le \tmplt{\Loss}{s-hinge}{GSC}(w;S) + \frac{4r\norm{w^*}}{\sqrt{m}} + (1+2r\norm{w^*})\sqrt{\frac{2ln(\frac{2}{\delta})}{m}}
\end{equation*}

Notice that $w^*$ is unknown. Therefore we show a data-dependant bound.
Define $\H^i = \{w: \norm{w} \le 2^i\}$ and $\delta_i = \frac{\delta}{2^i}$. Note that $\sum_{i=1}^{\infty} \delta_i = \delta$.
Similarly to the first part of the proof, for all $i$, $\forall w\in \H_i$ with probability greater or equal to $\delta$:
\begin{equation*}
    \tmplt{\Loss}{s-hinge}{GSC}(w;D) \le \tmplt{\Loss}{s-hinge}{GSC}(w;S) + \frac{4r\cdot2^i}{\sqrt{m}} + (1+2\cdot2^i)\sqrt{\frac{2ln(\frac{2}{\delta_i})}{m}}
\end{equation*}

From the union bound, we get that with probability greater or equal to $1-\delta$ this holds for all $\H_i$. This means that for all $w\in \H$ we have for
$i=\left \lceil \log(\norm{w}) \right \rceil \le \log(\norm{w}) + 1$:

\begin{equation*}
    \tmplt{\Loss}{s-hinge}{GSC}(w;D) \le \tmplt{\Loss}{s-hinge}{GSC}(w;S) + \frac{8r\norm{w}}{\sqrt{m}} + (1+4r\norm{w})\sqrt{\frac{2ln(\frac{4\norm{w}}{\delta})}{m}}
\end{equation*}

Plugging $w=w_S$ finishes the proof.
This proof can be adjusted easily to work for the soft formulation algorithm.
\end{proof}

\subsection{Tightened generalization bounds for notable subclasses of GSC}
\label{pr:bounds}
Let $D$ be a joint distribution over triples $(x,z,y)$,
and let $r=\max_{x \in \X} \norm{x}$.
Denote by $\what$ the minimizer of the empirical generalized strategic hinge loss with $L_2$ regularization.
Then for every $\delta \in [0,1]$,
if the training set $\smplst$ includes $m$ samples,
then w.p. $\ge 1-\delta$ it holds that:
\begin{equation*}
    \tmplt{\Loss}{0/1}{} \leq 
    \tmplt{\Lossapx}{s-hinge}{GP} +  \frac{4r\norm{\what}}{\sqrt{m}} 
    + (1+2\tmplt{\rho}{}{GP} \norm{w}) \sqrt{\frac{2 \ln(\frac{4\norm{\what}}{\delta})}{m}}
\end{equation*}
where $\tmplt{\Loss}{0/1}{}$ is the expected 0/1 loss,
$\tmplt{\Lossapx}{s-hinge}{GP}$ is the empirical s-hinge loss on $\smplst$,
and
$\tmplt{\rho}{}{GP} = r+2$.
\begin{proof}
This proof follows the proof of Theorem \ref{eq:bound_gsc} with two slight differences:

\textbf{Rademacher complexity of $\H_k\circ S$}: \\
\begin{align*}
&mR(\H_k\circ S) = \mathbb{E}_{\sigma}\left[\sup_{w\in \H_k}\sum_{i=1}^m \sigma_i y_{i} (w^\top x_{i} +2 z_i \norm{w}) \right] =_1 & \\
&\mathbb{E}_{\sigma}\left[\sup_{w\in \H_k}\sum_{i=1}^m \sigma_i(w^\top x_{i} +2 z_i \norm{w}) \right] =\mathbb{E}_{\sigma}\left[\sup_{w\in \H_k}\sum_{i=1}^m \sigma_i(w^\top x_{i} +2 z_i \norm{w}) \right] \le & \\
&\mathbb{E}_{\sigma}\left[\sup_{w\in \H_k}\sum_{i=1}^m \sigma_i w^\top x_{i} \right] + \mathbb{E}_{\sigma}\left[\sup_{w\in \H_k}\sum_{i=1}^m \sigma_i \cdot 2 z_i \norm{w} \right] =_2 &\\
&\mathbb{E}_{\sigma}\left[\sup_{w\in \H_k} w^\top \sum_{i=1}^m \sigma_i x_{i} \right] + \mathbb{E}_{\sigma}\left[\sup_{w\in \H_k} 2 \norm{w} \sum_{i=1}^m \sigma_i \right] = &\\
&\mathbb{E}_{\sigma}\left[\sup_{w\in \H_k} w^\top \sum_{i=1}^m \sigma_i x_{i} \right] \le_3  \mathbb{E}_{\sigma}\left[k \norm{\sum_{i=1}^m \sigma_i x_{i}} \right] = &\\
&k\mathbb{E}_{\sigma}\left[\sqrt{\norm{\sum_{i=1}^m \sigma_i x_{i}}^2 }\right] \le_4 
k\sqrt{\mathbb{E}_{\sigma}\left[\norm{\sum_{i=1}^m \sigma_i x_{i}}^2 \right]} = &\\
&k\sqrt{\mathbb{E}_{\sigma}\left[\sum_{i,j}^m \sigma_i \sigma_j \left \langle x_i, x_j  \right \rangle \right]} =_5 
k\sqrt{\sum_{i=1}^m\mathbb{E}_{\sigma}\left[\sigma_i^2\right]\norm{x_i}^2} \le & \\
&k\sqrt{m}\cdot \max_i\norm{x_i} \le k\sqrt{m}\cdot \max_i\norm{x_i} = k\sqrt{m}\cdot r&
\end{align*}
where steps follow from:
\begin{enumerate}
    \item $\sigma_i y_i \equiv_d \sigma_i$ under the expectancy
    \item $\sigma_i z_i \equiv_d \sigma_i$ under the expectancy
    \item Cauchy-Schwartz inequality
    \item Jensen's inequality
    \item Independence
\end{enumerate}    
Therefore, $R(\H_k\circ S) \le \frac{kr}{\sqrt{m}}$.

\textbf{Derivation of} $\tmplt{\rho}{}{NL}, \tmplt{\rho}{}{GP}, \tmplt{\rho}{}{ADV}, \tmplt{\rho}{}{SC}$: \\
The Rademacher complexity generalization theorem states that $c$ is equal to the maximal loss for any sample in $S$. Therefore for GP, ADV and SC:
\begin{align*}
    &\max_{s\in S} \tmplt{\Loss}{s-hinge}{}(s) = \max_{s\in S} \max\{0, 1-yw^\top x + 2yz\norm{w}\} = 1+(r+2)\norm{w} \Rightarrow&\\
    &\Rightarrow \tmplt{\rho}{}{GP}, \tmplt{\rho}{}{ADV}, \tmplt{\rho}{}{SC} = r+2&
\end{align*}

However, a close examination of the theorem proof \citep{RademacherProof} allows for the partition of the sample-set $S$ into several subsets $S_1, S_2, ... S_k$. Then $c = \sum_{i=1}^{k}\frac{\mathbb{E}_D[|S_i|]}{|S|}\max_{s\in S_i} loss(s)$.\\

In NL, the sample set can be partitioned into $2$ subsets:\\
$S_1 = \{s=(x, \ytilde, y)\in S: \ytilde=y\}, S_2 = \{s=(x, \ytilde, y)\in S: \ytilde\neq y\}$. \\
$\mathbb{E}_D[|S_1|] = 1-\epsilon, \mathbb{E}_D[|S_2|] = \epsilon$.\\
$\max_{s\in S_1} \tmplt{\Loss}{s-hinge}{NL}(s) = \max_{s\in S_1} \max\{0, 1-yw^\top x - 2\norm{w}\} = 1+(r-2)\norm{w}$. \\
$\max_{s\in S_2} \tmplt{\Loss}{s-hinge}{NL}(s) = \max_{s\in S_2} \max\{0, 1-yw^\top x + 2\norm{w}\} = 1+(r+2)\norm{w}$. \\

In conclusion, $c = \epsilon(1+(r+2)\norm{w}) + (1-\epsilon)(1+(r-2)\norm{w}) = 1+(r+2-4\epsilon)\norm{w}$.

\end{proof}
\endgroup

\end{document}